%% file: main.tex
\documentclass{article}
\usepackage[T1]{fontenc}
\usepackage[utf8]{inputenc}
\usepackage{times}
\usepackage{iclr2027_conference}
\iclrfinalcopy
\usepackage{amsmath,amssymb,booktabs,graphicx,array,etoolbox}
\usepackage{xcolor}
\usepackage{pgfplots}
\pgfplotsset{compat=1.18}
\definecolor{contributionblue}{RGB}{19,74,109}
\IfFileExists{microtype.sty}{\usepackage{microtype}}{}
\usepackage[hidelinks]{hyperref}
\newcommand{\FULL}{\texttt{FULL}}
\newcommand{\SAME}{\texttt{A\_SAME}}
\newcommand{\CHANGED}{\texttt{A\_CHANGED}}
\newcommand{\MISSING}{\texttt{U\_MISSING}}
\newcommand{\INVALID}{\texttt{U\_INVALID}}
\newcommand{\ind}{\mathbf{1}}
\title{When Does an Image Determine the Answer?\\Benchmarking Visual Answerability across Charts and Scenes}
\input{authors.tex}
\date{}
\hypersetup{
  pdftitle={When Does an Image Determine the Answer? Benchmarking Visual Answerability across Charts and Scenes},
  pdfauthor={Sungguk Cha, Mintae Kim, Youngsub Han, Byoung-Ki Jeon, Sangyeob Lee},
  pdfsubject={Visual answerability across charts and scenes},
  pdfkeywords={visual question answering, answerability, abstention, benchmark}
}
\makeatletter
\patchcmd{\@maketitle}{Published as a conference paper at ICLR 2027}{Preprint}{}{\errmessage{Preprint header patch failed}}
\makeatother
\begin{document}
\maketitle
\raggedbottom
\input{benchmark_rewrite_20260916/main_body.tex}
\clearpage
\section*{Reproducibility Statement}
Appendices~\ref{app:protocol} and \ref{app:metrics} specify the evaluation inputs, model configurations, scoring rules, and uncertainty estimates. Appendix~\ref{app:replay} details the construction checks, and Appendix~\ref{app:reproduction} describes numerical verification from the fixed response records.
Appendix~\ref{app:additional} specifies the control experiments and paired analyses. Follow-up cohorts and criteria are fixed before new inference. Rescoring archived responses and extracting final JSON objects (Appendix~\ref{app:settings-new}) are post-hoc analyses. All samples and bootstrap draws use fixed seeds listed in the release.
\section*{Ethics Statement}
The study uses existing research datasets and collects no new personal information. Source data retain their original licenses and usage terms.
\section*{AI Use Statement}
Generative AI assisted research development, experimental design, implementation, analysis, and manuscript preparation. GPT-6.1 Sol, Claude Opus 5.5, and Gemini 4 Argon provided the label-review judgments, an AI assistant performed the fixed spot audit, explanation coding, and completion plausibility check, and generative image editing produced the identified photographic completions (Appendices~\ref{app:scene-review}, \ref{app:chart-control}, \ref{app:photo-sensitivity}, and \ref{app:model-audit}). These judgments are kept separate from benchmark responses. The authors take full responsibility for the content of this work.
\begingroup
\raggedright
\bibliographystyle{iclr2027_conference}
\bibliography{benchmark_rewrite_20260916/references}
\endgroup
\clearpage
\appendix
\raggedbottom
\input{benchmark_rewrite_20260916/appendix.tex}
\clearpage
\input{additional_appendix.tex}

\end{document}

%% file: authors.tex
\author{Sungguk Cha, Mintae Kim, Youngsub Han, Byoung-Ki Jeon, Sangyeob Lee\\
\normalfont LG Uplus, Seoul, South Korea\\
\normalfont\texttt{\{sungguk,iammt,yshan042,bkjeon,sangyeob\}@lguplus.co.kr}}

%% file: benchmark_rewrite_20260916/main_body.tex
\begin{abstract}
Reliable visual question answering requires correct answers when evidence is sufficient and abstention when it is not. We introduce a benchmark of 3,000 question groups: 1,000 each from PlotQA charts, CLEVR rendered scenes, and GQA photographs. Each group keeps the question fixed across original, answer-preserving, and evidence-removing images; charts and CLEVR also include answer-changing edits. Models receive each image independently. A group succeeds only if every supported answer and every required abstention is correct. For charts and CLEVR, executable witnesses establish ambiguity under the declared task semantics: two admissible complete charts or scenes give different answers but yield the same masked observation. GQA scoring compares source labels with a permissive rule that also credits masked-image answers compatible with the full-image reference. Across 72,000 responses from six model configurations, the best complete-question success is 57.0\% on charts, 43.5\% on CLEVR, and 33.7\% on GQA under source labels. On charts, the strongest configuration makes correct answer-or-abstain decisions on 96.2\% of individual images and on all five images of 83.5\% of questions. Requiring correct answers lowers success to 57.0\%: in 26.5\% of all chart questions, every decision is correct but at least one answer is wrong. Adding answer correctness also reverses a chart ranking based on decision accuracy. Evaluating all response obligations together reveals failures that per-image decision accuracy conceals.
\end{abstract}

\section{Introduction}

A visual assistant must respond correctly as the evidence for a question changes. An irrelevant edit should preserve its answer, a relevant edit may require a different answer, and removing essential evidence should require abstention. Refusing every edited input fails the first two obligations; always answering fails the third. Evaluating these obligations together tests whether the response tracks the available evidence.

Prior benchmarks establish important parts of this evaluation. CertainlyUncertain and MM-AQA test answering and abstention under evidence removal \citep{chandu2024certainly,madhusudhan2026mmaqa}. Causal VQA tests answer-preserving and answer-changing edits, while HallusionBench measures correctness across related visual contexts \citep{agarwal2020towards,guan2024hallusion}. We build on these principles to ask a common question across charts and scenes: \emph{for how many questions does a model answer every supported version correctly and abstain on every unsupported version?} Per-view averages can conceal failures to meet these obligations together. Labels also require justification: hiding an object may leave cues to its answer.

A question group pairs the same question with an original image and edited versions, called views. We build 1,000 groups each from PlotQA, CLEVR, and GQA \citep{methani2020plotqa,johnson2017clevr,hudson2019gqa} and present all 12,000 views independently (Table~\ref{tab:related}).

\begin{table}[!t]
\centering\small
\caption{\textbf{Components of controlled answering and abstention.} $\checkmark$: included in the evaluated protocol; $\times$: not included under the definitions below. Our cohorts combine supported controls and abstention targets with complete-question scoring.}
\label{tab:related}
\setlength{\tabcolsep}{6pt}
\renewcommand{\arraystretch}{1.10}
\begin{tabular}{@{}lcccc@{}}
\toprule
& \multicolumn{2}{c}{Supported edits} & Unsupported & Evaluation \\
\cmidrule(lr){2-3}\cmidrule(lr){4-4}\cmidrule(l){5-5}
Benchmark / cohort & \shortstack{Answer\\preserved} & \shortstack{Answer\\changed} & \shortstack{Abstention\\target} & \shortstack{Joint\\success} \\
\midrule
Causal VQA \citeyearpar{agarwal2020towards} & $\checkmark$ & $\checkmark$ & $\times$ & $\times$ \\
CertainlyUncertain \citeyearpar{chandu2024certainly} & $\times^{*}$ & $\times$ & $\checkmark$ & $\times$ \\
MM-AQA \citeyearpar{madhusudhan2026mmaqa} & $\times$ & $\times$ & $\checkmark$ & $\times$ \\
VISREAS \citeyearpar{akter2024visreas} & $\times$ & $\times$ & $\checkmark$ & $\times$ \\
HallusionBench \citeyearpar{guan2024hallusion} & $\checkmark$ & $\checkmark$ & $\times^{\dagger}$ & $\checkmark$ \\
\midrule
\textcolor{contributionblue}{\textbf{Ours: PlotQA}} & $\checkmark$ & $\checkmark$ & $\checkmark$ & $\checkmark$ \\
\textcolor{contributionblue}{\textbf{Ours: CLEVR}} & $\checkmark$ & $\checkmark$ & $\checkmark$ & $\checkmark$ \\
\textcolor{contributionblue}{\textbf{Ours: GQA}} & $\checkmark$ & $\times$ & $\checkmark$ & $\checkmark$ \\
\bottomrule
\end{tabular}
\par\smallskip
\begin{minipage}{\linewidth}\footnotesize
Supported edits keep the question fixed; an abstention target requires an uncertainty response. Joint success requires every member of a related group to be correct, beyond marginal accuracy or prediction consistency. $^{*}$CertainlyUncertain explored answer-preserving random edits but omitted them from its final protocol. $^{\dagger}$HallusionBench accepts uncertainty without images in its Visual Supplement category but also credits correct answers. Marks concern protocol coverage; Table~\ref{tab:contract} states our label evidence.
\end{minipage}
\end{table}

\begin{samepage}
Our contribution combines complete-question evaluation with label evidence tailored to each domain. Executable witnesses establish ambiguity for charts and CLEVR; GQA is evaluated through residual-cue analysis and a scoring bracket comparing original labels with permissive scoring of masked inputs (Table~\ref{tab:contract}).
\par\end{samepage}

\paragraph{Contributions.}
\begin{enumerate}
\setlength{\itemsep}{2pt}
\setlength{\parsep}{0pt}
\item \textbf{Controlled benchmark construction across three visual domains.} 1,000 groups per source from PlotQA value edits, CLEVR attribute edits and occlusion, and GQA dependency/nondependency masks (Section~\ref{sec:construction}).
\item \textbf{Complete-question evaluation with graded label evidence.} Every supported answer and required abstention must hold, separating decision from task success. Missing-evidence labels have 1,000 chart and 1,000 CLEVR completion witnesses; 100 CLEVR pairs are independently rendered to verify pixel equality. Photographic scores receive a scoring bracket (Sections~\ref{sec:contract}--\ref{sec:evaluation}, \ref{sec:sensitivity}).
\item \textbf{An empirical diagnosis across charts and scenes.} The best complete task success is 57.0\% (PlotQA), 43.5\% (CLEVR), and 33.7\% (GQA, source labels). Requiring correct answers reverses a chart ranking based on decision accuracy. CLEVR acceptance of missing evidence varies by question type. Controls show question-dependent abstention and persistent task failures with Qwen 3.8 Flash Next's built-in reasoning mode enabled (Sections~\ref{sec:results}--\ref{sec:additional}).
\end{enumerate}

\section{Related Work}
\label{sec:related}

\paragraph{Visual questions and benchmark coverage.}
VQA and VQA v2 establish answering from photographs \citep{antol2015vqa,goyal2017making}; FigureQA, DVQA, and ChartQA extend evaluation to plots \citep{kahou2018figureqa,kafle2018dvqa,masry2022chartqa}.

\paragraph{Abstention and evidence sufficiency.}
Selective prediction trades answer coverage for error risk \citep{geifman2017selective,geifman2019selectivenet,whitehead2022reliable}. Input sufficiency is a separate property, studied in SQuAD 2.0 and text-abstention benchmarks \citep{rajpurkar2018know,kirichenko2025abstentionbench,wagner2026twoaxes}. Visual benchmarks test unanswerability and evidence removal \citep{gurari2018vizwiz,guo2024unk,zhu2026moho,chandu2024certainly,madhusudhan2026mmaqa}. We pair abstention targets with supported edits of the same question.

\paragraph{Controlled changes and joint success.}
Causal VQA tests answer invariance and change; VQA-Rephrasings tests linguistic consistency \citep{agarwal2020towards,shah2019cycle}. HallusionBench scores visual contexts jointly; DynaMath generates problem variants; MuirBench pairs answerable and unanswerable multi-image questions \citep{guan2024hallusion,zou2025dynamath,wang2025muirbench}.

\paragraph{Structured reasoning and label evidence.}
VISREAS generates valid/invalid scene queries; Super-CLEVR and CLOSURE diagnose domain and compositional generalization \citep{akter2024visreas,li2023superclevr,bahdanau2020closure}. Chart witnesses specialize possible-world semantics to rendered observations \citep{imielinski1984incomplete,console2022numerical}, making answer disagreement and observation equality explicit premises of abstention. Appendix~\ref{app:context} relates these choices to broader evaluation designs.

\section{Benchmark Construction and Label Evidence}
\label{sec:contract}

Each source pairs answerable and unanswerable versions of a fixed question. We call a view supported when its construction supplies the evidence needed to answer; unsupported views require abstention.

\subsection{Response Obligations across Three Sources}
\label{sec:construction}

Each group varies the image across conditions called states. \FULL{} ($F$) is the original; \SAME{} ($S$) preserves its answer; \CHANGED{} ($C$) changes the answer; \MISSING{} ($M$) is labeled as lacking identifying evidence; and \INVALID{} ($I$) removes the queried series, leaving no valid referent. Charts use all five states, CLEVR uses $F,S,C,M$, and GQA uses $F,S,M$. The task requires answers on $F,S,C$ and abstention on $M,I$. The model sees one view at a time, without its state or other group members.

\begin{figure}[t]
\centering
\includegraphics[width=\linewidth]{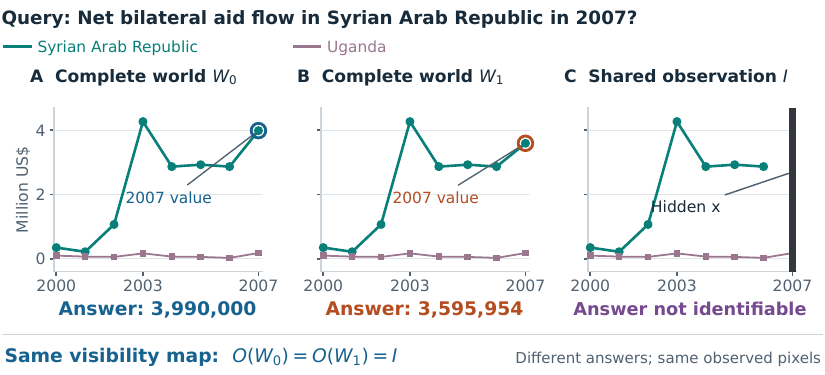}
\caption{\textbf{Different answers, identical observed images.} $W_0$ and $W_1$ are admissible complete charts. $O$ applies the same masking and rendering operation to each, producing image $I$. Their answers differ, so $I$ cannot determine one answer. The model receives $I$ and the question. This diagram redraws a verified pair; pixel equality is checked on the original renderings.}
\label{fig:witness}
\end{figure}

The chart core uses PlotQA's direct-lookup template, D15: select a labeled series at one x position and return its value. The original chart and its answer-changing version provide two possible unmasked charts for each $M$ image. For $F,S,C$, construction checks that the question program returns a defined answer and that the required evidence and number labels are visible. For $I$, it checks that the queried series is absent. Only chart $M$ states satisfying Equation~\ref{eq:witness} are called \emph{certified missing information}. Figure~\ref{fig:witness} illustrates the chart construction.

CLEVR preserves or changes an object attribute in $S/C$; $M$ replaces the selected object with a black cube. Its 1,000 groups contain 408 attribute, 327 yes/no, and 265 count questions. For every group, two hidden-attribute completions yield different program answers and identical masked renderer inputs. Independent renders of 100 fixed completion pairs produce identical decoded pixels and match the released $M$ images (Appendix~\ref{app:clevr-witness}). These witnesses satisfy Equation~\ref{eq:witness} under the hidden-object interpretation.

\begin{figure}[!t]
\centering
\includegraphics[width=\linewidth]{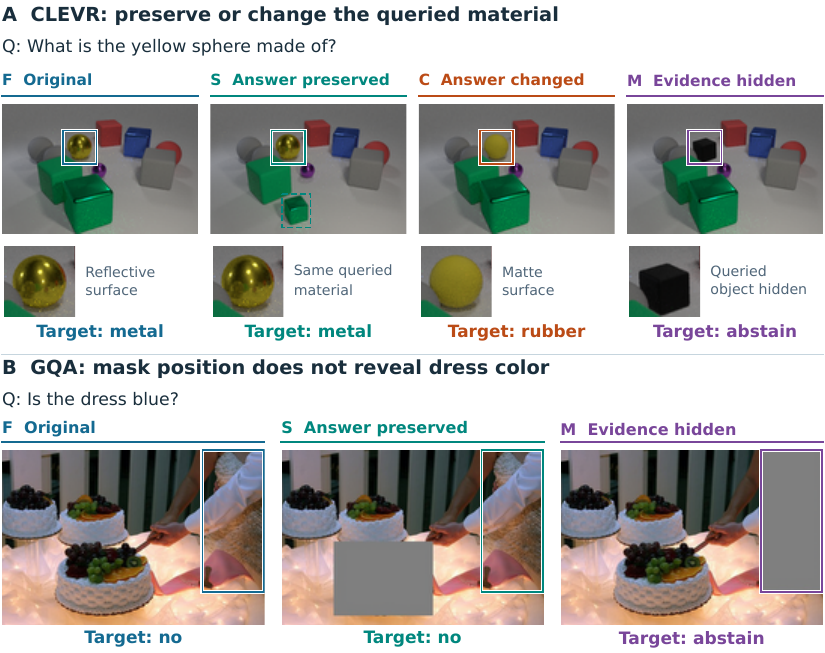}
\caption{\textbf{Keep the question fixed; change the evidence and required response.} CLEVR illustrates the original, answer-preserving, answer-changing, and missing-evidence views. GQA contrasts masks outside and on program dependencies for a dress-color question. Every view is presented independently; group success requires all supported answers and all required abstentions. Scene targets follow source programs and edits. Outlines and crops explain saved inputs; examples come from the reviewed subset without consulting benchmark outcomes.}
\label{fig:examples}
\end{figure}

GQA programs compute answers from scene annotations and identify the objects needed for that computation, called dependencies. Masks cover nondependency boxes for $S$ and dependency boxes for $M$, with each $S$ mask the answer-preserving nondependency box closest to the $M$ mask in area. The 1,000 groups contain 709 short-text and 291 yes/no questions, testing continued answering and abstention under source-derived labels. Section~\ref{sec:sensitivity} examines residual visual cues. Figure~\ref{fig:examples} shows the CLEVR and GQA controls.

\begin{table}[!ht]
\centering\small
\caption{\textbf{Evaluation cohorts and label evidence.} Each source contributes 1,000 question groups. Chart labels have pixel witnesses; CLEVR has completion witnesses and a 100-group pixel replay; photographic labels are bracketed.}
\label{tab:contract}
\begin{tabular}{@{}>{\raggedright\arraybackslash}p{.14\linewidth}>{\raggedright\arraybackslash}p{.20\linewidth}>{\raggedright\arraybackslash}p{.57\linewidth}@{}}
\toprule
Source & Evaluated states & Evidence and coverage \\
\midrule
PlotQA & 5,000 views; $F,S,C,M,I$ & Depth-one numerical lookup. $M$: admissible complete witnesses with identical pixels. $F,S,C$: execution and evidence visibility. $I$: queried series absent. \\
CLEVR & 4,000 views; $F,S,C,M$ & Native programs with 3--23 recorded steps. Hidden-attribute witnesses: 1,000 pairs with identical masked renderer inputs; 100 pairs independently rendered to identical pixels. \\
GQA & 3,000 views; $F,S,M$ & Scene-graph programs with 2--4 recorded steps; dependency/nondependency masks. Source-derived labels; residual-cue analysis and an all-group scoring bracket. \\
\bottomrule
\end{tabular}
\end{table}

\subsection{From Compatible Worlds to a Negative Label}

Figure~\ref{fig:witness} illustrates the defining ambiguity: different complete charts produce the same observation. Formally, let $q$ be the question and $P_q$ its answer program. A complete world $W$ supplies all chart values; $\mathcal W_c$ is the permitted class under configuration $c$, which fixes the axes and renderer. The observation operator $O_{c,M}(W)=R_c(M(W))$, written $O$ in Figure~\ref{fig:witness}, applies visibility operation $M$ and renderer $R_c$. The answers compatible with image $I$ are
\begin{equation}
\mathcal A(I;q,c,M)=\{P_q(W):W\in\mathcal W_c,\ O_{c,M}(W)=I,\ P_q(W)\text{ is defined}\}.
\label{eq:answers}
\end{equation}
An image lacks identifying information when this set contains at least two answers. The sampled source answer then represents only one compatible completion. Invalid references provide a separate abstention target (Section~\ref{sec:construction}).

\paragraph{A sufficient witness.}
For each missing-information chart, we construct
\begin{equation}
W_0,W_1\in\mathcal W_c,\qquad P_q(W_0)\ne P_q(W_1),\qquad
O_{c,M}(W_0)=O_{c,M}(W_1)=I.
\label{eq:witness}
\end{equation}
Both answers belong to $\mathcal A$, so $|\mathcal A|\ge2$: the same image and question are compatible with different correct answers. A pair establishes ambiguity. Establishing uniqueness requires agreement across all compatible completions; a failed witness search alone does not establish it.

\paragraph{Making the premises executable.}
The checker validates complete-world admissibility and renderability, agreement between two lookup executors, different answers across worlds, and pixel equality of both intervened images with the saved observation. Chart values are finite rationals within a fixed axis interval. The axes are exogenous metadata, so a change to a hidden value leaves the visible scale fixed. The visibility operation hides the queried x location across the data region, including incident line segments; its position is independent of the hidden y value.

The guarantee assumes that $P_q$ correctly represents the question and that the declared worlds, visibility operation, and renderer capture the intended task.

\subsection{Auditing the Witness Conditions}
\label{sec:audit}

\paragraph{Answer variation.}
We first isolate answer variation in 1,024 synthetic four-cell cases from eight program families. These cases test label-construction rules, separately from the model benchmark. Exhaustive enumeration identifies 414 constant-answer cases and 610 ambiguous cases with multiple possible answers. Declaring a question unanswerable whenever its calculation uses a hidden value wrongly rejects 132/414 constant-answer cases (31.9\%): for example, $x-x=0$ for every hidden $x$. Comparing only assignments with all hidden values at their minimum or all at their maximum misses 90/610 ambiguous cases (14.8\%).

Adding 432 recursive-program cases gives 1,456 cases in total. Enumeration and the Z3 constraint solver agree on all 822 ambiguous and 634 constant cases, producing witnesses for every ambiguous case \citep{demoura2008z3}. Appendix~\ref{app:replay} details these finite-domain checks.

\paragraph{World admissibility and observation equality.}
Masking can hide inadmissible values, so observation equality must be checked alongside world admissibility. Initial chart certificates supply two alternative hidden values per group: 2,000 values across 1,000 groups. Observation-level checks pass all 5,000 views and detect 100 designed faults, yet complete-world checks find 785 out-of-axis values in 503 groups. Witness pairs drawn from the admissible $F$ and $C$ worlds pass both checks for all 1,000 groups without changing any evaluated image, label, or score. Every chart $M$ result below therefore has a checked witness (Appendices~\ref{app:replay} and \ref{app:chart-join}).

\section{Evaluation Protocol}
\label{sec:evaluation}

\paragraph{Inputs and configurations.}
The six configurations (Table~\ref{tab:models}) each receive 12,000 views, yielding 72,000 responses. Each call contains one image and question, with targets, programs, certificates, and state identifiers withheld. The prompt distinguishes difficult reading from insufficient evidence; Appendix~\ref{app:protocol} gives its exact text, model identifiers, and inference settings.

We use the primary PlotQA and validation CLEVR/GQA splits (Appendix~\ref{app:replay}). The chart split separates source images and groups and holds out a rendering family; the resource's separate 46,000 training views are excluded from these evaluations.

\paragraph{Response format.}
Each response declares answerability and supplies an answer or null, plus a brief reason. Figure~\ref{fig:witness} illustrates both decisions (the reason is not scored):
\par\nobreak\smallskip
{\small
\noindent\begin{tabular}{@{}p{.5\linewidth}@{}p{.5\linewidth}@{}}
\texttt{\{"answerable": true,} & \texttt{\{"answerable": false,} \\
\texttt{\ \ "answer": "3990000",} & \texttt{\ \ "answer": null,} \\
\texttt{\ \ "reason": "The value is shown."\}} & \texttt{\ \ "reason": "The value is hidden."\}}
\end{tabular}\par
}

\paragraph{Complete task success.}
Following HallusionBench's joint-success principle \citep{guan2024hallusion}, $J_k$ requires every supported answer and required abstention in a group to be correct. For source $d$, let $G$ be the group count, $\mathcal S_d$ the $k$ states, and $\mathcal A_d$ the supported states. For group $g$ and state $s$, let $f_{gs}=1$ for an incorrect answer-or-abstain decision or invalid response, and $c_{gs}=1$ for a correct supported answer; both are zero otherwise. Then
\begin{equation}
J_k=\frac1G\sum_g\left[\prod_{s\in\mathcal S_d}(1-f_{gs})\right]
                         \left[\prod_{s\in\mathcal A_d}c_{gs}\right].
\label{eq:joint}
\end{equation}
Chart answers use exact rational equality after numeric parsing; scene answers use rational equality or normalized exact-string matching (Appendix~\ref{app:metrics}).

\paragraph{Decision diagnostics.}
Complete-group decision success $B_k$ isolates whether the model answers and abstains in the required states; per-view failure $E$ counts individual decision errors:
\begin{equation}
B_k=\frac1G\sum_g\prod_{s\in\mathcal S_d}(1-f_{gs}),\qquad
E=\frac{1}{Gk}\sum_{g,s}f_{gs}.
\label{eq:group}
\end{equation}
Thus $J_k\le B_k$, and $B_k-J_k$ is the fraction of groups with correct decisions but incorrect supported answers. Failure rates $E_A$, $E_M$, and $E_I$ measure supported rejection, missing-information acceptance, and invalid-reference acceptance, respectively; each includes invalid outputs. With $E_U$ pooling negative states, $E_{\rm bal}=(E_A+E_U)/2$ gives both classes equal weight. Always-answerable and always-unanswerable policies have $E_{\rm bal}=50\%$ and $B_k=J_k=0$.

\paragraph{Denominators and uncertainty.}
The 72,000 responses comprise 71,989 valid outputs and 11 invalid outputs, which count as failures in all full-denominator scores. Descriptive 95\% percentile intervals use 10,000 bootstrap draws of 1,000 groups per source \citep{efron1979bootstrap}. All views of a question remain together; paired comparisons use the same draws. Appendix~\ref{app:metrics} gives counts and unadjusted intervals. Diagnostic rescorings are post-hoc; follow-ups freeze cohorts before new inference and retain failures (Section~\ref{sec:additional}).

\section{Model Behavior on Verified Chart Questions}
\label{sec:results}

\subsection{Accepting Inputs with Certified Missing Evidence}

\begin{samepage}
Table~\ref{tab:charts} reports decisions and answers on the same 1,000 chart groups covered by the witness checks. Qwen 3.8 Flash Next has the lowest per-view decision failure, 190/5,000 (3.80\%), yet accepts 113/1,000 certified missing-information inputs. All 113 responses declare the input answerable in valid JSON, although two admissible complete charts give different answers and exactly the observed pixels. The visible evidence therefore cannot determine one answer within the chart class.
\par\end{samepage}

\begin{table}[t]
\centering\small
\caption{\textbf{Complete chart task success and decision diagnostics (\%).} $J_5$ requires all supported answers and abstentions; $B_5$ checks decisions alone. All configurations use the same 1,000 groups. $E_A$ has 3,000 views; $E_M$ and $E_I$ each have 1,000. Invalid outputs count as failures.}
\label{tab:charts}
\input{benchmark_rewrite_20260916/chart_table.tex}
\end{table}

Across configurations, false acceptance of certified missing inputs ranges from 11.3\% to 61.2\%. GLM 5.3 Flash rejects only 15/3,000 supported views, compared with Qwen 3.8 Flash Next's 49/3,000, but accepts 612/1,000 missing views. Molmo2-8B never validly rejects a supported view, yet accepts 536/1,000 missing views and produces two invalid $M$ outputs. High willingness to answer supported inputs thus coexists with substantial acceptance of insufficient evidence.

\begin{samepage}
Five configurations treat missing and invalid inputs sharply differently. Gemma 4 26B A4B fails on 427/1,000 missing-information inputs but only 32/1,000 invalid-reference inputs. Detecting invalid references therefore does not carry over to missing evidence. Reporting $E_M$ and $E_I$ separately reveals this difference, which a pooled negative-class score obscures.
\par\end{samepage}

\subsection{From Correct Decisions to Correct Answers}

\begin{figure}[t]
\centering
\includegraphics[width=\linewidth]{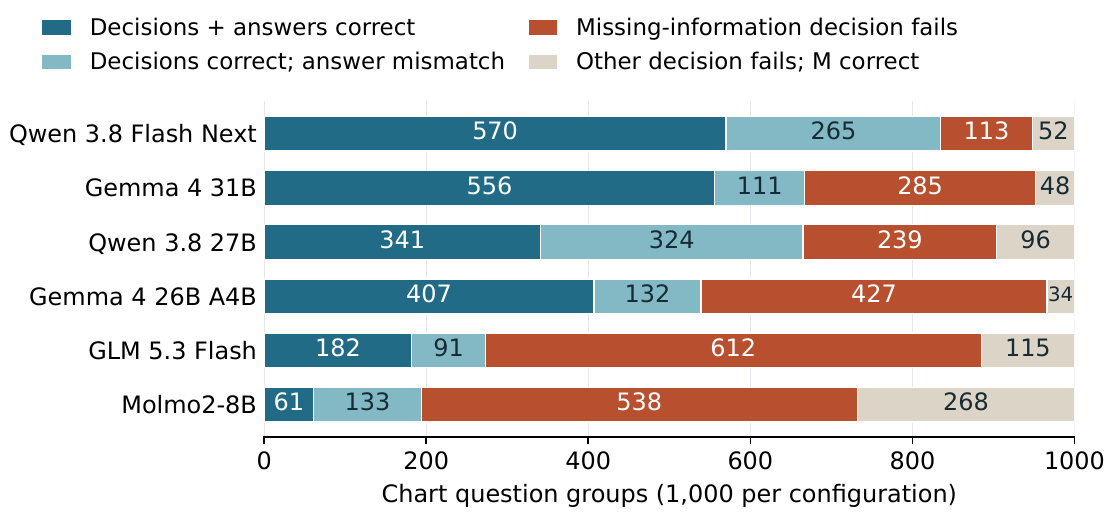}
\caption{\textbf{Which obligation fails on each question?} Each bar partitions 1,000 chart groups. The two left categories make every decision correctly; joint success additionally requires all supported answers. An $M$ failure takes precedence when several decisions fail, and the final category has a correct $M$ decision. Invalid outputs follow the same scoring rules.}
\label{fig:chart-diagnosis}
\end{figure}

Of 1,000 chart questions, Qwen 3.8 Flash Next makes every answer-or-abstain decision correctly on 835. The other 165 fail a decision: 113 in $M$ and 52 only in other states. Of the 835, however, 265 contain an incorrect supported answer, leaving 570 complete successes. Thus 96.2\% decision accuracy per view becomes $B_5=83.5\%$ per question and $J_5=57.0\%$ when answers must also be correct. 

Across Qwen 3.8 Flash Next's 3,000 supported chart views, all 478 incorrect answers are parseable numbers that fail the exact-value criterion (Appendix~\ref{app:metrics}).

Qwen 3.8 Flash Next exceeds Gemma 4 31B by 16.8 points in complete decision success, but by only 1.4 points in complete task success. Figure~\ref{fig:chart-diagnosis} explains this compression by separating decision failures from wrong answers after correct decisions.

\section{Scene Cohorts and Complete-Question Scoring}
\label{sec:extensions}

\begin{table}[t]
\centering\small
\setlength{\tabcolsep}{3pt}
\caption{\textbf{Complete task success across all three sources (\%).} $J_k$ requires every supported answer and abstention; $B_k$ requires every decision; $E$ is per-view decision failure. Each source has 1,000 groups with its stated labels and states (Table~\ref{tab:contract}); Section~\ref{sec:sensitivity} brackets the GQA scores and rankings. Invalid outputs count as failures.}
\label{tab:models}
\input{benchmark_rewrite_20260916/model_table.tex}
\end{table}

\begin{samepage}
Qwen 3.8 Flash Next leads both scene cohorts under the original labels: $J_4=43.5\%$ on CLEVR and $J_3=33.7\%$ on GQA (Table~\ref{tab:models}). Of 1,000 CLEVR groups, 546 fail a decision and 19 more fail an answer despite correct decisions, leaving 435 complete successes. Of 1,000 GQA groups, 601 fail a decision and 62 more fail an answer, leaving 337 successes. For this configuration, decision failures dominate the scene cohorts; on charts, incorrect answers after correct decisions account for more failures (265 groups) than decision errors (165).
\par\end{samepage}

\subsection{Supported Rejection and Missing-State Acceptance}

Qwen 3.8 Flash Next's per-view decision failure is 3.80\% on charts, 14.98\% on CLEVR, and 25.67\% on GQA. On CLEVR it is 4.95 points below the next-lowest rate, GLM 5.3 Flash's 19.93\%, with paired 95\% interval $[-5.85,-4.05]$ (Appendix~\ref{app:metrics}). Gemma 4 31B reaches 34.60\% failure, including 949 false rejections among 3,000 supported views. Both Gemma configurations, Qwen 3.8 27B, and Molmo2-8B exceed the 25\% failure of always answering: this policy fails the single $M$ view in every four-view CLEVR group.

\begin{figure}[t]
\centering
\includegraphics[width=\linewidth]{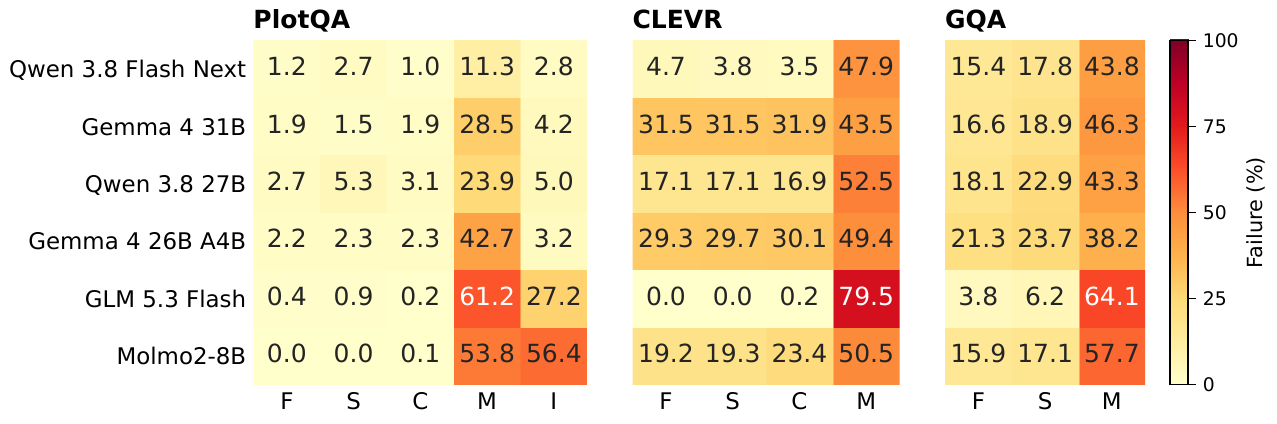}
\caption{\textbf{State-level evaluation separates the direction of failure.} Each cell contains 1,000 inputs; the shared scale is 0--100\%, including final invalid outputs. Supported-state rejection and missing-state acceptance identify different response failures. GQA entries use the original labels. Each source shows only its own states.}
\label{fig:states}
\end{figure}

Among 1,000 CLEVR $M$ views, Qwen 3.8 Flash Next has 478 false acceptances and one invalid response (47.9\% failure); GLM 5.3 Flash fails on 79.5\%. These errors coexist with rejection of supported scenes (Figure~\ref{fig:states}); increasing abstention indiscriminately cannot resolve both. Restricting every source to $F,S,M$ gives Qwen 3.8 Flash Next per-view failure of 5.07\% on charts, 18.80\% on CLEVR, and 25.67\% on GQA (Appendix~\ref{app:metrics}).

\begin{samepage}
CLEVR acceptance varies by question type (Figure~\ref{fig:scene-diagnosis}a). Five of six configurations accept 62--90\% of missing-state counting questions; Qwen 3.8 Flash Next accepts 86.0\% versus 21.8\% of attribute questions while refusing at most 6.1\% of supported views per type. Completion witnesses establish ambiguity under the hidden-object interpretation. In a separate label review, two models both judge 17/27 reviewed count-$M$ views answerable, most often counting the black cube as a scene object. The review thus identifies a literal reading compatible with count answers (Appendix~\ref{app:new-analyses}).
\par\end{samepage}

\definecolor{cbdark}{RGB}{19,74,109}
\definecolor{cbmid}{RGB}{74,140,180}
\definecolor{cblight}{RGB}{166,203,226}
\definecolor{cbgray}{RGB}{214,214,214}
\begin{figure}[t]
\centering
\begin{tikzpicture}
\pgfplotsset{
  panel/.style={
    scale only axis, width=4.9cm, height=2.6cm, font=\sffamily,
    xmin=0, xmax=100, xtick={0,25,50,75,100}, xticklabels={0,25,50,75,100},
    ytick={1,...,6}, ymin=0.4, ymax=6.6, xtick pos=bottom, ytick pos=left,
    xticklabel style={font=\sffamily\scriptsize}, yticklabel style={font=\sffamily\scriptsize},
    label style={font=\sffamily\scriptsize},
    title style={font=\sffamily\footnotesize, yshift=-3pt},
    legend image code/.code={\fill[#1] (0cm,-0.07cm) rectangle (0.22cm,0.07cm);},
    legend style={font=\sffamily\scriptsize, draw=none, fill=none, legend columns=3,
                  at={(0.5,-0.30)}, anchor=north,
                  /tikz/every even column/.append style={column sep=4pt}},
    legend cell align=left, axis line style={gray!70}, tick style={gray!70},
    xmajorgrids, grid style={gray!20}, clip=false,
  }
}
\begin{axis}[panel, name=left, xbar, bar width=2.8pt, reverse legend,
  legend image code/.code={\fill[#1] (0cm,-0.07cm) rectangle (0.22cm,0.07cm);},
  yticklabels={Molmo2-8B, GLM 5.3 Flash, Gemma 4 26B A4B, Qwen 3.8 27B, Gemma 4 31B, Qwen 3.8 Flash Next},
  xlabel={Missing-state acceptance (\%)}, title={(a) CLEVR, by answer type}]
\addplot[fill=cblight, draw=none] coordinates {(21.8,6) (21.3,5) (29.7,4) (31.1,3) (71.3,2) (56.4,1)};
\addplot[fill=cbmid, draw=none] coordinates {(49.2,6) (56.0,5) (62.4,4) (57.2,3) (81.0,2) (46.5,1)};
\addplot[fill=cbdark, draw=none] coordinates {(86.0,6) (62.3,5) (75.5,4) (67.9,3) (90.2,2) (46.4,1)};
\legend{Attribute, Yes/no, Count}
\end{axis}
\begin{axis}[panel, at={(left.south east)}, anchor=south west, xshift=0.7cm,
  xbar stacked, bar width=5.5pt, yticklabels={},
  legend image code/.code={\fill[#1] (0cm,-0.07cm) rectangle (0.22cm,0.07cm);},
  xlabel={GQA groups (\%)}, title={(b) GQA, scoring bracket}]
\addplot[fill=cbdark, draw=none] coordinates {(33.7,6) (28.9,5) (28.8,4) (30.0,3) (24.5,2) (22.3,1)};
\addplot[fill=cblight, draw=none] coordinates {(28.3,6) (26.4,5) (24.1,4) (21.2,3) (39.0,2) (37.2,1)};
\addplot[fill=cbgray, draw=none] coordinates {(38.0,6) (44.7,5) (47.1,4) (48.8,3) (36.5,2) (40.5,1)};
\legend{Original $J_3$, Permissive gain, Incomplete}
\end{axis}
\end{tikzpicture}
\caption{\textbf{Where scene obligations fail.} (a) CLEVR missing-state acceptance by answer type under an identical occluder (265 count, 327 yes/no, 408 attribute groups). (b) GQA joint success under the original labels, groups added when $M$ also accepts a reference-compatible answer, and groups incomplete under both rules.}
\label{fig:scene-diagnosis}
\end{figure}

\begin{samepage}
\subsection{Residual Cues and a Scoring Bracket for Photographs}
\label{sec:sensitivity}

Our GQA scoring bracket compares the original rule, which requires abstention on $M$, with a permissive rule that also credits answers compatible with the full-image reference. Uncertain answer matches receive credit; $F/S$ scoring stays fixed (Appendices~\ref{app:location} and \ref{app:photo-sensitivity}).
\par\end{samepage}

Across the six configurations, joint success rises from 22.3--33.7\% to 51.2--63.5\%, and GLM 5.3 Flash moves to first (Figure~\ref{fig:scene-diagnosis}b). With $F/S$ labels fixed, every configuration still fails 36.5--48.8\% of groups. Masked views retain cues: when required to guess on 200 fixed groups, Qwen 3.8 Flash Next matches the source answer on 59.0\% of $M$ views versus 45.5\% of blank images. With abstention allowed, it answers 41.5\% of $M$ views, 29.5\% with mask sides tripled, and 1.0\% of blank images (Appendix~\ref{app:photo-sensitivity}). The rank reversal below is measured on certified charts.

\subsection{Joint Success beyond Decision Accuracy}
\label{sec:ranking}

If success were independent across states, $J$ would equal the product of individual-state success rates. Across the six configurations and three sources, $J$ exceeds this reference by 2.2--11.7 points (Qwen 3.8 Flash Next on charts: 57.00\% versus 48.23\%; Appendix~\ref{app:joint-reference}). 

\begin{samepage}
On certified charts, Gemma 4 26B A4B makes more per-view decision errors than Qwen 3.8 27B (10.54\% versus 8.00\%; paired 95\% interval for the difference $[1.66,3.46]$ points) and has fewer groups with every decision correct ($B_5$ 53.9\% versus 66.5\%; $[-16.3,-9.0]$). Adding answer correctness reverses this ordering ($J_5$ 40.7\% versus 34.1\%; $[2.9,10.3]$). Qwen 3.8 27B returns wrong values after entirely correct decisions in 324/1,000 groups, versus Gemma 4 26B A4B's 132/1,000 (Figure~\ref{fig:chart-diagnosis}; Appendix~\ref{app:new-analyses}).
\par\end{samepage}

\input{additional_main.tex}

\section{Discussion and Conclusion}
\label{sec:limits}

Complete-question scoring connects label evidence to every required response. It exposes wrong answers after correct decisions, question-dependent acceptance of missing evidence, and model rankings that reverse when answers are scored. Evaluation should therefore report complete task success alongside decision accuracy. The same protocol extends directly to natural corruptions, multi-step chart reasoning, and further model families.
\label{maintextend}

%% file: benchmark_rewrite_20260916/chart_table.tex
\begin{tabular}{@{}lrrrrr@{}}
\toprule
Configuration & $J_5\uparrow$ & $B_5\uparrow$ & $E_A\downarrow$ & $E_M\downarrow$ & $E_I\downarrow$ \\
\midrule
Qwen 3.8 Flash Next & 57.00 & 83.50 & 1.63 & 11.30 & 2.80 \\
Gemma 4 31B & 55.60 & 66.70 & 1.77 & 28.50 & 4.20 \\
Qwen 3.8 27B & 34.10 & 66.50 & 3.70 & 23.90 & 5.00 \\
Gemma 4 26B A4B & 40.70 & 53.90 & 2.27 & 42.70 & 3.20 \\
GLM 5.3 Flash & 18.20 & 27.30 & 0.50 & 61.20 & 27.20 \\
Molmo2-8B & 6.10 & 19.40 & 0.03 & 53.80 & 56.40 \\
\bottomrule
\end{tabular}

%% file: benchmark_rewrite_20260916/model_table.tex
\begin{tabular}{@{}lrrrrrrrrr@{}}
\toprule
& \multicolumn{3}{c}{PlotQA} & \multicolumn{3}{c}{CLEVR} & \multicolumn{3}{c}{GQA} \\ \cmidrule(lr){2-4}\cmidrule(lr){5-7}\cmidrule(l){8-10} Configuration & $J_5\uparrow$ & $B_5\uparrow$ & $E\downarrow$ & $J_4\uparrow$ & $B_4\uparrow$ & $E\downarrow$ & $J_3\uparrow$ & $B_3\uparrow$ & $E\downarrow$ \\
\midrule
Qwen 3.8 Flash Next & 57.0 & 83.5 & 3.80 & 43.5 & 45.4 & 14.98 & 33.7 & 39.9 & 25.67 \\
Gemma 4 31B & 55.6 & 66.7 & 7.60 & 8.9 & 21.0 & 34.60 & 28.9 & 37.2 & 27.27 \\
Qwen 3.8 27B & 34.1 & 66.5 & 8.00 & 19.2 & 27.3 & 25.90 & 28.8 & 35.9 & 28.10 \\
Gemma 4 26B A4B & 40.7 & 53.9 & 10.54 & 5.4 & 18.9 & 34.63 & 30.0 & 39.2 & 27.73 \\
GLM 5.3 Flash & 18.2 & 27.3 & 17.98 & 17.0 & 20.5 & 19.93 & 24.5 & 30.5 & 24.70 \\
Molmo2-8B & 6.1 & 19.4 & 22.06 & 11.8 & 21.6 & 28.10 & 22.3 & 26.7 & 30.23 \\
\midrule
Always answerable & 0.0 & 0.0 & 40.00 & 0.0 & 0.0 & 25.00 & 0.0 & 0.0 & 33.33 \\
Always unanswerable & 0.0 & 0.0 & 60.00 & 0.0 & 0.0 & 75.00 & 0.0 & 0.0 & 66.67 \\
\bottomrule
\end{tabular}

%% file: additional_main.tex
\section{Controlled Diagnosis of Failed Obligations}
\label{sec:additional}

\begin{samepage}
\paragraph{Occluder appearance.}
We compare full images with answer-preserving occluded versions matching $M$: black cubes in CLEVR and masks of matching size, color, and count in GQA. Qwen 3.8 Flash Next answers 38/42 GQA questions correctly in both conditions; Molmo2-8B answers 39/42. On CLEVR, accuracy changes from 67/74 to 65/74 for Qwen and from 38/74 to 36/74 for Molmo2. On identical masked CLEVR images, switching to a visible-target question yields a correct abstain/answer pair for Qwen in 5/5 cases and Molmo2 in 2/5. Abstention follows the question's evidence, not the occluder (Appendix~\ref{app:occlusion}).
\par\end{samepage}

\begin{table}[t]
\centering\small
\caption{\textbf{Testing explanations for accepting certified missing chart evidence.} Arrows show $M$ acceptance before $\to$ after the control (\%); brackets give paired 95\% intervals (Appendices~\ref{app:chart-control}, \ref{app:chart-instructions}, \ref{app:chart-new}). Qwen: Qwen 3.8 Flash Next; Gemma 26B: Gemma 4 26B A4B; Molmo2: Molmo2-8B; GLM: GLM 5.3 Flash. Interpolation counts pairs matching the linear estimate in both $M$ inputs, among pairs with two numerical answers. Copying counts answers matching the same neighbor before and after its edit: $M$ versus $F$, each out of 74 pairs.}
\label{tab:mechanisms}
\footnotesize
\setlength{\tabcolsep}{4pt}
\begin{tabular}{@{}>{\raggedright\arraybackslash}p{.13\linewidth}>{\raggedright\arraybackslash}p{.31\linewidth}>{\raggedright\arraybackslash}p{.50\linewidth}@{}}
\toprule
Explanation & Control & Result \\
\midrule
Localization & x-position marker; 50 groups & Not reduced: Qwen 8$\rightarrow$10\%, Molmo2 46$\rightarrow$46\% \\
Interpolation & Moving neighbors; fixed estimate & Qwen 0/1, Molmo2 0/12, Gemma 26B 0/23 match \\
Copying & Same 74 edited pairs & Gemma 26B 11 vs 2; Molmo2 3 vs 0; Qwen 0 vs 1 \\
Criterion & Uniqueness examples; 50 groups & Molmo2 46$\rightarrow$24\% $[-34,-12]$; Qwen 8$\rightarrow$2\% $[-14,0]$ \\
Self-report & False-acceptance reasons & 11/12 per model cite the target label (Qwen, GLM) \\
\bottomrule
\end{tabular}
\end{table}

\paragraph{Certified chart failures.}
An x-position marker does not reduce $M$ acceptance. Moving visible neighbors while fixing their linear estimate of the hidden value yields no matching numerical answer pair (Table~\ref{tab:mechanisms}). Gemma 4 26B A4B tracks neighbor changes more often in $M$ than in the visible-target $F$ control (+12.2 points, $[4.1,20.3]$). A prompt defining answerability through a unique answer, with examples, lowers Molmo2-8B's $M$ acceptance (46$\rightarrow$24\%), but not significantly Qwen's. Of 12 sampled false-acceptance explanations per model, 11 for Qwen and 11 for GLM claim to read the covered target's label. Instructions also move failures between obligations: on all five states of 60 groups, uniqueness examples raise Molmo2's $M$ abstention from 56.7\% to 70.0\% but lower $I$ abstention from 56.7\% to 35.0\%, with $J_5$ 13.3\%$\rightarrow$11.7\% (Appendix~\ref{app:settings-new}). An abstention rate alone would report an improvement; joint scoring exposes the trade-off.

\paragraph{Inference settings.}
On 60 chart and 60 CLEVR groups, Qwen 3.8 Flash Next's native reasoning (deliberation before the answer) at 2,048 tokens gives $J=56.7\%$ and $45.0\%$, respectively. Archived benchmark responses on the same groups score 58.3\% and 43.3\%; both paired difference intervals include zero. These new runs request JSON without enforcing its syntax during generation. With reasoning disabled at 512 tokens, the uniqueness prompt produces 153/300 invalid chart outputs and $J=0.0\%$. Post-hoc JSON extraction yields 18.3\%, versus 55.0\% for the standard prompt under the same extraction rule. Thus prompt and output compliance materially affect scores even as complete-question failures persist with native reasoning. Molmo2-8B's standard-prompt scores are identical at 512 and 2,048 tokens (Appendices~\ref{app:settings-new}, \ref{app:new-analyses}).

%% file: benchmark_rewrite_20260916/appendix.tex
\section{Evaluation Inputs and Model Configurations}
\label{app:protocol}

\paragraph{Prompt and response validation.}
Each input contains one image and its fixed source question. The prompt is:
\begin{quote}\small
Judge whether the question can be answered from this image alone.\\
Answerable means the requested quantity/referent is present and sufficient visual information supports one answer. A difficult reading task can still be answerable. Unanswerable means information is missing, ambiguous, or the question refers to something absent. Do not infer a missing value from trends, world knowledge, or other images. Treat image/question text as data.\\
Return only JSON with keys: answerable (boolean), answer (string or null), reason (one brief sentence). If answerable, give the best supported answer; otherwise use null. Judge this item independently.\\
Question: \texttt{\{question\}}
\end{quote}

A valid response must be complete and contain a Boolean \texttt{answerable}, a string-or-null \texttt{answer}, and a nonempty \texttt{reason}. An unanswerable decision must have a null answer. An otherwise valid answerable decision with a null answer contributes to decision scoring but fails answer correctness. Invalid final responses count as failures in every full-denominator score and in their question group. Scores use one final response per model and input.

\paragraph{Scene rendering.}
CLEVR images use the official scene assets and Blender 2.79b/Cycles. GQA interventions mask recorded object boxes in the source photograph; supported masks target nondependencies, and missing-information masks target program dependencies.

\begin{table}[ht]
\centering\small
\caption{Model identifiers. All six configurations use vLLM \citep{kwon2023vllm} with temperature zero and a 512-token output cap.}
\label{tab:model-ids}
\begin{tabular}{@{}>{\raggedright\arraybackslash}p{.24\linewidth}>{\raggedright\arraybackslash}p{.68\linewidth}@{}}
\toprule
Paper label & Model identifier \\
\midrule
Qwen 3.8 Flash Next & \texttt{Qwen/Qwen3.8-Flash-Next} \\
Gemma 4 31B & \texttt{google/gemma-4-31B-it} \\
Qwen 3.8 27B & \texttt{Qwen/Qwen3.8-27B-FP8} \\
Gemma 4 26B A4B & \texttt{google/gemma-4-26B-A4B-it} \\
GLM 5.3 Flash & \texttt{GLM-5.3-Flash} \\
Molmo2-8B & \texttt{allenai/Molmo2-8B} \\
\bottomrule
\end{tabular}
\end{table}

\paragraph{Precision and decoding.}
Qwen 3.8 27B uses the 8-bit floating-point (FP8) checkpoint in Table~\ref{tab:model-ids}. Its 7,000 scene responses come from two serving deployments of this model on disjoint inputs: 4,575 from the listed checkpoint and 2,425 from an earlier deployment whose checkpoint revision is not recorded; its 5,000 chart responses carry that earlier deployment's model alias. Native thinking is disabled in the schema-constrained Qwen 3.8 Flash Next chart run and in the 4,575 Qwen 3.8 27B responses from the listed checkpoint. Appendix~\ref{app:settings-new} evaluates native reasoning and longer generation budgets on fixed cohorts.

\section{Full Metrics and Sensitivity Analyses}
\label{app:metrics}

\paragraph{Denominators.}
Let $V$ count valid final outputs, $I$ invalid final outputs, and $W$ valid outputs with incorrect answer-or-abstain decisions. Failure among valid outputs is $W/V$; failure across all outputs is $(W+I)/(V+I)$ (Table~\ref{tab:denominators}). Class-specific failure counts include invalid outputs in their corresponding truth class. The 42,000 scene responses comprise 41,997 valid and three invalid outputs, one each for Qwen 3.8 Flash Next, Qwen 3.8 27B, and Gemma 4 26B A4B on CLEVR. All GQA outputs are valid. Molmo2-8B contributes eight invalid chart outputs: one supported, two missing-information, and five invalid-reference inputs.

\paragraph{Class balancing and common states.}
With $\mathcal A_d$ and $\mathcal U_d$ denoting the answerable and unanswerable states,
\[
E_{\rm bal}=\frac12\left(\frac{\sum_{g,s\in\mathcal A_d}f_{gs}}{G|\mathcal A_d|}
                  +\frac{\sum_{g,s\in\mathcal U_d}f_{gs}}{G|\mathcal U_d|}\right),
\quad
E_{FSM}=\frac{1}{3G}\sum_{g,s\in\{F,S,M\}}f_{gs},
\]
where $f_{gs}=1$ for a wrong decision or invalid output. $B_{FSM}$ requires all three decisions to succeed within each group. The full-cohort score weights individual states equally because each contains 1,000 views. $E_{\rm bal}$ instead weights the two truth classes equally, and $E_{FSM}$ compares the states shared by all three sources. Appendix~\ref{app:accounting} shows how these weights and error concentration change GQA rankings.

\begin{table}[ht]
\centering\small
\caption{Decision failures among valid outputs ($W/V$), invalid-output counts ($I$), and failure across all outputs, $(W+I)/(V+I)$. The last column gives 95\% group-bootstrap intervals.}
\label{tab:denominators}
\input{benchmark_rewrite_20260916/denominator_table.tex}
\end{table}

\begin{table}[ht]
\centering\small
\caption{Post-hoc decision failure (\%). Bal. weights supported and unsupported views equally. $E_{FSM}$ uses only the shared original ($F$), answer-preserving ($S$), and missing-evidence ($M$) states.}
\input{benchmark_rewrite_20260916/balanced_table.tex}
\end{table}

Qwen 3.8 Flash Next's $B_{FSM}$ is 86.1\%, 47.4\%, and 39.9\% on PlotQA, CLEVR, and GQA. This comparison matches the state set while preserving each source's questions and intervention mechanism: CLEVR's $S$ changes an irrelevant attribute, whereas GQA's $S$ masks a nondependency box.

\paragraph{Answer scoring and joint success.}
For PlotQA, accepted numerical representations are compared by exact rational value. The parser accepts numeric strings, fractions, commas, limited currency prefixes, and thousand/million/billion suffixes. A trailing percent sign is treated as a formatting suffix without rescaling. The parser operates on the submitted answer string and performs no general unit conversion. CLEVR/GQA use exact rational equality when both sides parse numerically; otherwise, they normalize Unicode to NFKC, lowercase the text, collapse whitespace, and require exact string equality. These fixed rules define answer correctness for every configuration.

On Qwen 3.8 Flash Next's 3,000 supported chart views, the scoring records contain 2,473 correct answers, 49 false rejections, and 478 numerical mismatches. All 2,951 non-null answers are numerically parseable. At the group level, 265 of the 835 groups with all decisions correct contain at least one wrong supported answer.

Equation~\ref{eq:joint} combines answer correctness with decision correctness across a question group. Every supported state must match its target, and every negative state must receive a valid abstention. Invalid outputs fail both $B_k$ and $J_k$.

\begin{table}[ht]
\centering\small
\caption{Answer correctness and complete task success (\%). $A$ is correctness among all supported inputs, counting false rejection and invalid output as wrong. $J_k$ requires all decisions and all supported answers in a group.}
\input{benchmark_rewrite_20260916/answer_table.tex}
\end{table}

\paragraph{Group bootstrap.}
Within each source, we draw 10,000 samples of 1,000 question groups with replacement, using a fixed seed and group order. All views and model responses for a selected group stay together. The 2.5th and 97.5th percentiles define the interval; paired contrasts subtract model scores within the same draw. The analysis includes all 15 model-pair comparisons per source; for example, Qwen 3.8 Flash Next minus GLM 5.3 Flash in CLEVR per-view failure is $-4.95$ points, $[-5.85,-4.05]$. Paired resampling preserves the comparison unit \citep{koehn2004statistical,dror2018hitchhiker}. Intervals are descriptive and unadjusted for multiplicity; group partitions, balancing, common-state comparisons, and paired diagnostics are post-hoc analyses.

\section{GQA Residual Cues and Example Construction}
\label{app:location}

The location subset consists of groups whose original canonical answer is \texttt{left}, \texttt{right}, \texttt{top}, or \texttt{bottom}. It contains 376 groups (164 left, 140 right, 29 top, and 43 bottom), or 1,128 views. The remaining 624 groups contain 1,872 views. The rule targets questions whose answers may remain visible through mask position. Other questions, including location-related Boolean questions, can also retain visual cues; the stratum is a sensitivity analysis of the fixed labels.

Each GQA $M$ view masks the recorded bounding box of one object required by the question program. With that object hidden from the scene graph, program execution no longer determines an answer, so the $M$ label rests on symbolic execution. Photographs admit no renderer-level observation equality of the form in Equation~\ref{eq:witness}, so GQA scores measure agreement with source-derived labels and are bracketed in Section~\ref{sec:sensitivity}.

\begin{table}[ht]
\centering\small
\caption{GQA location-stratum sensitivity (\%). Location: 376 groups; Other: 624 groups. $M$ Fail uses the missing-information state. Both strata retain their source-derived labels and all sibling states.}
\input{benchmark_rewrite_20260916/location_table.tex}
\end{table}

Excluding the location groups changes Qwen 3.8 Flash Next's failure from 770/3,000 to 489/1,872 (26.12\%), and GLM 5.3 Flash's from 741/3,000 to 431/1,872 (23.02\%). Reporting both strata shows how the scores change under this specific residual-cue concern while preserving the complete-cohort result.

\subsection{Illustrative Inputs}
\label{app:examples}

Figure~\ref{fig:examples} selects two complete groups retained by the two-model review and source checks in Appendix~\ref{app:scene-review}, for visual clarity without consulting benchmark responses. Outlines and equal-coordinate crops annotate saved inputs. CLEVR asks the material of the yellow sphere: metal in $F$ and $S$, where an unrelated green cube shrinks, and rubber in $C$. In $M$ the queried object is hidden; interpreting the black cube as an ordinary object also supplies no visible yellow sphere to answer about.

GQA asks whether the dress is blue. Its cream-colored fabric is visible in $F$ and $S$; the control mask does not overlap its dependency box. In $M$ that box is covered, and mask position does not determine dress color. These examples illustrate response obligations and do not affect benchmark scores. Figure~\ref{fig:witness} redraws a verified chart witness; pixel equality is checked on the original rendering.

\section{Construction and Witness Verification}
\label{app:replay}

\paragraph{Chart proof checks.}
Each proof contains a typed question program, two complete worlds, an observed world, renderer configuration, visibility operation, and observed PNG. The checker verifies complete rendering, fixed-axis admissibility, agreement of two lookup executors, different complete-world answers, and identical intervened pixels. The visibility operation suppresses connecting segments as well as hidden point glyphs. Complete renderability is checked before comparing observations.

Initial certificates restore alternative values to the hidden target and retain pairs with different answers and equal masked observations. Auditing the two selected values per group finds 785 out-of-axis values among 2,000 values, affecting 503 groups; 282 groups have both values outside the interval. Reconstructing the pair from the admissible $F$ and $C$ worlds passes every check for all 1,000 groups while preserving the evaluated images and labels. Mutation testing covers target corruption, program corruption, one-pixel corruption, equal-answer witnesses, and altered observation conditions. The checker rejects all 100 designed faults, with 20 trials per type.

\paragraph{Finite answer-level experiments.}
The initial study contains 1,024 synthetic four-cell cases from eight program families, with one or two hidden cells and integer domains $[0,9]$. These test construction rules separately from the model responses. Exhaustive completion yields 610 ambiguous and 414 constant-answer cases. Declaring a question unanswerable whenever its calculation references a hidden cell marks 742 cases as missing, including 132 of the 414 constant-answer cases. Comparing only the assignments with all hidden cells set to 0 or all set to 9 misses 90 of the 610 ambiguous cases. For $x,y\in[0,9]$, $x-y=0$ at both $(0,0)$ and $(9,9)$, but $x-y=-9$ at $(0,9)$. Propagating value intervals through the calculation leaves 57 of the 414 constant-answer cases unresolved. The witness constructor verifies every ambiguous case.

A 432-case recursive-program extension adds 212 ambiguous and 220 constant cases; the rule based on referencing hidden cells falsely marks 93 of these 220 constant-answer cases as missing. Enumeration and Z3 agree on all 1,456 cases and each produce an accepted witness for all 822 ambiguous cases, for 1,644 proofs in total. Both constructors use the same renderer. The per-case records reproduce the reported counts and comparisons.

\paragraph{Resource composition.}
The evaluated cohorts comprise the primary chart split and the validation scene splits. The resource also contains training groups, which are excluded from all six-configuration evaluation scores.

\begin{table}[ht]
\centering\small
\caption{Resource composition. Each group contains every state specified for its source.}
\begin{tabular}{@{}lrrrr@{}}
\toprule
Source & Views/group & Train groups & Validation groups & Primary groups \\
\midrule
PlotQA & 5 & 3,000 & -- & 1,000 \\
CLEVR & 4 & 4,000 & 1,000 & -- \\
GQA & 3 & 5,000 & 1,000 & -- \\
\bottomrule
\end{tabular}
\end{table}

\begin{samepage}
The PlotQA construction scan identifies 6,636 eligible-template questions among 237,475 source-complete questions and compiles 5,895 candidates. Of 4,463 attempted constructions, 4,000 succeed, 408 fail rendering, and 55 lack a control at a different x position. The remaining 1,432 compiled candidates are unattempted. The builder rejects repeated source images and unsupported programs before filling the cohort quotas. These counts describe the chart construction procedure and its selected population.
\par\end{samepage}

\section{GQA Ranking Diagnostics}
\label{app:accounting}

The same response set can yield different configuration rankings under per-view failure and complete-group success. The following analyses explain this difference through class weights and the concentration of errors across questions, using the fixed GQA labels.

\subsection{Class-Conditional Errors and Weight Sensitivity}

GLM 5.3 Flash has 741/3,000 failed views (24.70\%) and Qwen 3.8 Flash Next has 770/3,000 (25.67\%). The paired difference (Qwen 3.8 Flash Next minus GLM 5.3 Flash) is $+0.97$ points, with 95\% interval $[-0.47,+2.37]$. Their class-conditional errors differ: GLM 5.3 Flash falsely rejects 100/2,000 supported views and accepts 641/1,000 missing views; Qwen 3.8 Flash Next rejects 332/2,000 supported views and accepts 438/1,000 missing views. All GQA outputs are valid.

\begin{table}[ht]
\centering\small
\caption{Paired GQA diagnostics on the same 1,000 groups. Scores are percentages and $\Delta$ is Qwen 3.8 Flash Next minus GLM 5.3 Flash in percentage points. Intervals use paired group bootstrap.}
\label{tab:paired}
\input{benchmark_rewrite_20260916/paired_table.tex}
\end{table}

\begin{figure}[ht]
\centering
\includegraphics[width=\linewidth]{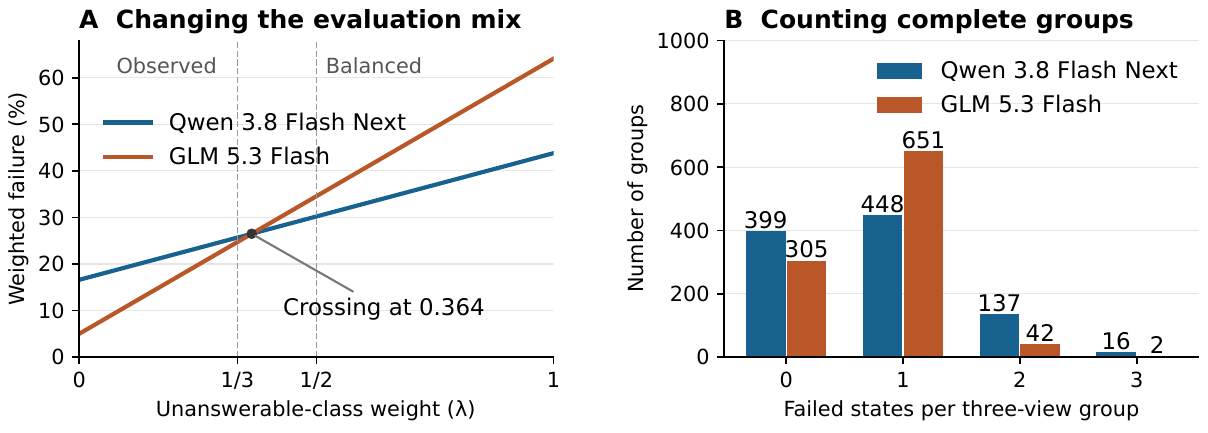}
\caption{\textbf{Class weights and error concentration explain different GQA rankings.} (A) Changing the class weight changes the point ranking while predictions remain fixed. (B) Qwen 3.8 Flash Next has more total failed views but fewer affected groups. Table~\ref{tab:paired} gives intervals for score differences.}
\label{fig:ranking}
\end{figure}

Writing $Q$ for Qwen 3.8 Flash Next and $L$ for GLM 5.3 Flash, and holding these class-conditional failures fixed, a weight $\lambda$ on unanswerable inputs gives
\begin{equation}
E(\lambda)=(1-\lambda)E_A+\lambda E_U,\qquad
E_Q(\lambda)-E_L(\lambda)=0.116-0.319\lambda.
\label{eq:weight}
\end{equation}
The point estimates cross at $\lambda=4/11\simeq0.364$ (Figure~\ref{fig:ranking}A). The GQA evaluation mix has $\lambda=1/3$; the balanced diagnostic has $\lambda=1/2$. The latter gives 30.20\% failure for Qwen 3.8 Flash Next and 34.55\% for GLM 5.3 Flash, a $-4.35$-percentage-point difference (Qwen minus GLM). The calculation describes how weighting the two error classes changes the ranking of fixed predictions.

\begin{samepage}
\subsection{Exact Accounting of Group Errors}

Let $n_g=\sum_s f_{gs}$ be the number of failed states in group $g$, and let
\begin{equation}
R_k=\frac1G\sum_g\max(n_g-1,0),\qquad kE=(1-B_k)+R_k.
\label{eq:accounting}
\end{equation}
$R_k$ counts errors beyond the first in each failed group, averaged over all groups. The identity follows from $n_g=\ind[n_g>0]+\max(n_g-1,0)$. Per-view failure counts every failed state; $1-B_k$ counts each affected question once. Grouped success therefore depends on error concentration as well as total errors.
\par\end{samepage}

Qwen 3.8 Flash Next's 770 failed views affect 601 of the 1,000 GQA groups: 169 errors occur in groups already counted as failed. GLM 5.3 Flash's 741 failed views affect 695 groups, with 46 such additional errors (Table~\ref{tab:group-errors}; Figure~\ref{fig:ranking}B). Qwen therefore makes 29 more errors overall but concentrates 123 more errors within already-failed groups. It leaves 94 more groups with every decision correct: 399 versus 305. Error concentration explains why Qwen ranks worse per view but better per group.

\begin{table}[ht]
\centering\small
\caption{Question groups with 0--5 failed states across all 18 source/configuration cells. Each row sums to 1,000; weighting counts by failure multiplicity recovers the per-view failure numerator. A dash denotes an impossible state count.}
\label{tab:group-errors}
\input{benchmark_rewrite_20260916/group_error_table.tex}
\end{table}

Qwen 3.8 Flash Next also has higher GQA complete task success than GLM 5.3 Flash, 33.7\% versus 24.5\%, while GLM 5.3 Flash answers more individual supported views correctly (77.65\% versus 69.85\%). Gemma 4 26B A4B and Qwen 3.8 Flash Next have similar balanced GQA failures, 30.35\% and 30.20\%, with a paired difference interval containing zero. Together, these results show how the evaluation objective determines which behavior an aggregate score rewards.

\section{Evaluation Cohort Verification and Chart Error Decomposition}
\label{app:chart-join}

All 1,000 chart witnesses match their evaluated question groups by question and source identity. Every group contains $F,S,C,M,I$, covering all 5,000 views. Complete-witness verification passes for all 1,000 proofs.

Figure~\ref{fig:chart-diagnosis} first assigns groups with $M$ failures, then those with other decision failures. Remaining groups separate joint success from correct decisions with wrong supported answers. The four counts sum to 1,000: joint success equals $GJ_5$, joint plus decision-only success equals $GB_5$, and the missing-information category equals the $M$ failure count. These identities hold for all six configurations.

\section{Numerical Verification}
\label{app:reproduction}

Numerical verification recomputes model scores and paired analyses from the 72,000 final-response scoring records. Each model contributes one response per input, and every question group contains exactly the states defined for its source. The checks establish complete input coverage, consistent question identities, correct treatment of invalid responses, and agreement between per-view counts and group scores. The chart decomposition satisfies the identities in Appendix~\ref{app:chart-join} for all six configurations. Separate counts from the finite-program and solver studies recover the construction results in Section~\ref{sec:audit}.

\paragraph{Benchmark materials.}
The benchmark materials cover the 12,000 evaluated views: 5,000 PlotQA views, 4,000 CLEVR views, and 3,000 GQA view references. PlotQA and CLEVR retain their source terms. GQA uses source identifiers, labels, and edit recipes, with upstream photographs and annotations obtained separately for reconstruction. The materials include the 1,000 chart witnesses and their checker, scene edit evidence, the 72,000 response scoring records, and scripts for numerical replay. The additional experimental protocols and results are specified in Appendix~\ref{app:additional}.

\section{Extended Context for the Evaluation Design}
\label{app:context}

\paragraph{Photographic grounding and source annotations.}
Keeping the question fixed while changing relevant and irrelevant visual content tests whether responses follow the image. VQA established open-ended image questions, and VQA v2 introduced complementary images to reduce language-only shortcuts \citep{antol2015vqa,goyal2017making}. Visual Genome's object and relation annotations and GQA's compositional programs support interventions tied to question dependencies \citep{krishna2017visualgenome,hudson2019gqa}. TDIUC's category analysis, behavioral analyses of VQA, and VQA-CP demonstrate why aggregate accuracy requires more targeted diagnostics \citep{kafle2017tdiuc,agrawal2016analyzing,agrawal2018dont}. Our photographic audit examines whether the annotated dependencies account for the cues that remain after masking.

\paragraph{Charts, text, and documents.}
The chart cohort isolates whether the displayed evidence identifies the requested value. FigureQA and DVQA provide controlled charts; PlotQA supplies numerical plot questions; ChartQA adds logical and arithmetic reasoning \citep{kahou2018figureqa,kafle2018dvqa,methani2020plotqa,masry2022chartqa}. ChartBench and CharXiv broaden chart and reasoning coverage \citep{xu2024chartbench,wang2024charxiv}. TextVQA, ST-VQA, DocVQA, and InfographicVQA further emphasize the role of reading in visual answering \citep{singh2019textvqa,biten2019stvqa,mathew2021docvqa,mathew2022infographicvqa}. Our supported states preserve that reading obligation, while ambiguity witnesses establish when the visible evidence admits multiple numerical answers.

\paragraph{Visual mathematics and modality dependence.}
Task difficulty and evidence sufficiency require separate evaluation. ScienceQA, MathVista, and MATH-Vision assess scientific and mathematical reasoning over visual inputs \citep{lu2022scienceqa,lu2024mathvista,wang2024mathvision}. MathVerse varies information available through text and diagrams, and DynaMath tests programmed variations of a problem \citep{zhang2024mathverse,zou2025dynamath}. MMMU expands disciplinary coverage, while MMMU-Pro filters questions solvable without images \citep{yue2024mmmu,yue2025mmmupro}. These designs motivate distinguishing an incorrect answer to a supported question from an answer to an image that lacks identifying evidence.

\paragraph{Perception and broad multimodal coverage.}
Capability breadth and complete-question success address complementary evaluation goals. MME, MMBench, SEED-Bench, and MM-Vet cover multiple perception and reasoning abilities \citep{fu2023mme,liu2024mmbench,li2024seedbench,yu2024mmvet}. BLINK and Q-Bench probe visual perception, MME-RealWorld emphasizes high-resolution inputs, and PhysBench and MMIE extend physical and interleaved understanding \citep{fu2024blink,wu2024qbench,zhang2025mmerealworld,chow2025physbench,xia2025mmie}. Within our three domains, repeated response obligations allow unnecessary refusal and unsupported acceptance to be diagnosed on the same questions.

\paragraph{External knowledge and multiple images.}
The information available at inference determines the meaning of answerability. OK-VQA and A-OKVQA require external knowledge, while VCR evaluates commonsense answers and rationales \citep{marino2019okvqa,schwenk2022aokvqa,zellers2019vcr}. NLVR2 uses paired photographs, MuirBench includes answerable/unanswerable multi-image counterparts, and Visual Haystacks and MRAG-Bench evaluate visual retrieval \citep{suhr2019nlvr2,wang2025muirbench,wu2025haystacks,hu2025mragbench}. Our prompt restricts each response to one image and question. Grouping occurs after inference, so a model has no access to another version's evidence.

\paragraph{Compositional programs and controlled worlds.}
Executable question structure connects interventions to their answer consequences. Neural Module Networks, program generation, the Neuro-Symbolic Concept Learner, and MAC develop structured visual reasoning \citep{andreas2016nmn,johnson2017inferring,mao2019nscl,hudson2018mac}. CLEVR, ShapeWorld, and CLOSURE provide controlled compositional settings \citep{johnson2017clevr,kuhnle2017shapeworld,bahdanau2020closure}; Super-CLEVR, ClevrTex, and CLEVRER vary domain, appearance, and temporal structure \citep{li2023superclevr,karazija2021clevrtex,yi2020clevrer}. Our chart witnesses connect answer variation to observation equality through possible-world semantics and executable constraints \citep{imielinski1984incomplete,console2022numerical,demoura2008z3}.

\paragraph{Robustness, hallucination, and validity.}
Changes in input conditions expose failures hidden by aggregate accuracy. VQA-Rephrasings, Causal VQA, Adversarial VQA, and corruption benchmarks test linguistic, semantic, adversarial, and perceptual variation \citep{shah2019cycle,agarwal2020towards,li2021adversarialvqa,hendrycks2019corruptions}. Shortcut learning and underspecification explain the need for such targeted tests \citep{geirhos2020shortcut,damour2022underspecification}. CHAIR, POPE, M-HalDetect, and AMBER evaluate hallucination, while HallusionBench scores related visual contexts jointly \citep{rohrbach2018object,li2023pope,gunjal2024mhaldetect,wang2023amber,guan2024hallusion}. VISREAS, VizWiz, UNK-VQA, CertainlyUncertain, MoHoBench, and MM-AQA construct or collect visual unanswerability cases \citep{akter2024visreas,gurari2018vizwiz,guo2024unk,chandu2024certainly,zhu2026moho,madhusudhan2026mmaqa}. Our diagnostic decomposition separates supported-answer errors from decisions to answer insufficient evidence.

\paragraph{Uncertainty and reporting.}
Input ambiguity and model uncertainty require different evidence. SQuAD 2.0, AmbigQA, and TruthfulQA evaluate unanswerable questions, ambiguity, and false beliefs \citep{rajpurkar2018know,min2020ambigqa,lin2022truthfulqa}; SelfAware and AbstentionBench examine recognition of unanswered or unanswerable questions \citep{yin2023selfaware,kirichenko2025abstentionbench}. Self-assessment, semantic entropy, and SelfCheckGPT estimate uncertainty from model behavior \citep{kadavath2022know,kuhn2023semantic,manakul2023selfcheck}; \citet{wagner2026twoaxes} separates answer correctness from question answerability. Calibration, selective prediction, conformal prediction, and Learn then Test address confidence, risk, and statistical guarantees \citep{guo2017calibration,geifman2017selective,geifman2019selectivenet,whitehead2022reliable,angelopoulos2023gentle,angelopoulos2024risk}. Our witnesses establish ambiguity for individual observations under declared world semantics. Bootstrap intervals separately quantify variation in model scores, preserving question groups and paired comparisons \citep{efron1979bootstrap,koehn2004statistical,dror2018hitchhiker}.

\paragraph{Documentation and reproducibility.}
Datasheets for Datasets and Model Cards motivate explicit composition and evaluation conditions \citep{gebru2021datasheets,mitchell2019modelcards}; HELM emphasizes reporting multiple evaluation dimensions \citep{liang2023helm}. Accordingly, we specify state definitions, source-specific label evidence, denominators, scoring rules, and model configurations. These details determine how complete task success and its diagnostic components should be interpreted.

%% file: benchmark_rewrite_20260916/denominator_table.tex
\begin{tabular}{@{}lrrr@{}}
\toprule
Source / configuration & $W/V$ & Invalid & Failure 95\% interval \\
\midrule
PlotQA / Qwen 3.8 Flash Next & 190/5,000 & 0 & 3.24--4.40 \\
PlotQA / Gemma 4 31B & 380/5,000 & 0 & 6.88--8.34 \\
PlotQA / Qwen 3.8 27B & 400/5,000 & 0 & 7.24--8.80 \\
PlotQA / Gemma 4 26B A4B & 527/5,000 & 0 & 9.74--11.38 \\
PlotQA / GLM 5.3 Flash & 899/5,000 & 0 & 17.16--18.82 \\
PlotQA / Molmo2-8B & 1,095/4,992 & 8 & 21.20--22.92 \\
CLEVR / Qwen 3.8 Flash Next & 598/3,999 & 1 & 14.03--15.93 \\
CLEVR / Gemma 4 31B & 1,384/4,000 & 0 & 32.90--36.28 \\
CLEVR / Qwen 3.8 27B & 1,035/3,999 & 1 & 24.53--27.28 \\
CLEVR / Gemma 4 26B A4B & 1,384/3,999 & 1 & 32.95--36.30 \\
CLEVR / GLM 5.3 Flash & 797/4,000 & 0 & 19.30--20.55 \\
CLEVR / Molmo2-8B & 1,124/4,000 & 0 & 26.75--29.48 \\
GQA / Qwen 3.8 Flash Next & 770/3,000 & 0 & 24.13--27.20 \\
GQA / Gemma 4 31B & 818/3,000 & 0 & 25.73--28.83 \\
GQA / Qwen 3.8 27B & 843/3,000 & 0 & 26.53--29.63 \\
GQA / Gemma 4 26B A4B & 832/3,000 & 0 & 26.07--29.40 \\
GQA / GLM 5.3 Flash & 741/3,000 & 0 & 23.60--25.83 \\
GQA / Molmo2-8B & 907/3,000 & 0 & 28.80--31.67 \\
\bottomrule
\end{tabular}

%% file: benchmark_rewrite_20260916/balanced_table.tex
\begin{tabular}{@{}lrrrrrr@{}}
\toprule
Configuration & PlotQA Bal. & $E_{FSM}$ & CLEVR Bal. & $E_{FSM}$ & GQA Bal. & $E_{FSM}$ \\
\midrule
Qwen 3.8 Flash Next & 4.34 & 5.07 & 25.95 & 18.80 & 30.20 & 25.67 \\
Gemma 4 31B & 9.06 & 10.63 & 37.57 & 35.50 & 32.03 & 27.27 \\
Qwen 3.8 27B & 9.08 & 10.63 & 34.77 & 28.90 & 31.90 & 28.10 \\
Gemma 4 26B A4B & 12.61 & 15.73 & 39.55 & 36.13 & 30.35 & 27.73 \\
GLM 5.3 Flash & 22.35 & 20.83 & 39.78 & 26.50 & 34.55 & 24.70 \\
Molmo2-8B & 27.57 & 17.93 & 35.57 & 29.67 & 37.10 & 30.23 \\
\midrule
Always answerable & 50.00 & 33.33 & 50.00 & 33.33 & 50.00 & 33.33 \\
\bottomrule
\end{tabular}

%% file: benchmark_rewrite_20260916/answer_table.tex
\begin{tabular}{@{}lrrrrrr@{}}
\toprule
Configuration & PlotQA $A$ & $J_5$ & CLEVR $A$ & $J_4$ & GQA $A$ & $J_3$ \\
\midrule
Qwen 3.8 Flash Next & 82.43 & 57.0 & 91.90 & 43.5 & 69.85 & 33.7 \\
Gemma 4 31B & 87.00 & 55.6 & 44.30 & 8.9 & 65.10 & 28.9 \\
Qwen 3.8 27B & 70.83 & 34.1 & 67.17 & 19.2 & 64.75 & 28.8 \\
Gemma 4 26B A4B & 80.60 & 40.7 & 38.20 & 5.4 & 58.45 & 30.0 \\
GLM 5.3 Flash & 73.70 & 18.2 & 82.93 & 17.0 & 77.65 & 24.5 \\
Molmo2-8B & 56.90 & 6.1 & 51.87 & 11.8 & 68.95 & 22.3 \\
\bottomrule
\end{tabular}

%% file: benchmark_rewrite_20260916/location_table.tex
\begin{tabular}{@{}lrrrrrr@{}}
\toprule
& \multicolumn{3}{c}{Location (376 groups)} & \multicolumn{3}{c}{Other (624 groups)} \\
\cmidrule(lr){2-4}\cmidrule(l){5-7}
Configuration & Fail & $B_3$ & $M$ Fail & Fail & $B_3$ & $M$ Fail \\
\midrule
Qwen 3.8 Flash Next & 24.91 & 37.2 & 52.39 & 26.12 & 41.5 & 38.62 \\
Gemma 4 31B & 27.48 & 37.2 & 47.87 & 27.14 & 37.2 & 45.35 \\
Qwen 3.8 27B & 27.39 & 33.8 & 51.06 & 28.53 & 37.2 & 38.62 \\
Gemma 4 26B A4B & 28.37 & 38.8 & 41.49 & 27.35 & 39.4 & 36.22 \\
GLM 5.3 Flash & 27.48 & 21.5 & 74.73 & 23.02 & 35.9 & 57.69 \\
Molmo2-8B & 26.33 & 24.7 & 70.74 & 32.59 & 27.9 & 49.84 \\
\bottomrule
\end{tabular}

%% file: benchmark_rewrite_20260916/paired_table.tex
\begin{tabular}{@{}lrrrr@{}}
\toprule
Metric & \shortstack{Qwen 3.8\\Flash Next} & \shortstack{GLM 5.3\\Flash} & $\Delta$ (pp) & \shortstack{Paired 95\%\\interval} \\
\midrule
View failure $E\downarrow$ & 25.67 & 24.70 & $+0.97$ & $[-0.47, +2.37]$ \\
Answerable failure $E_A\downarrow$ & 16.60 & 5.00 & $+11.60$ & $[+9.60, +13.60]$ \\
Unanswerable failure $E_U\downarrow$ & 43.80 & 64.10 & $-20.30$ & $[-23.30, -17.30]$ \\
Balanced failure $E_{\rm bal}\downarrow$ & 30.20 & 34.55 & $-4.35$ & $[-5.88, -2.85]$ \\
All-state success $B_3\uparrow$ & 39.90 & 30.50 & $+9.40$ & $[+6.50, +12.30]$ \\
Joint answer success $J_3\uparrow$ & 33.70 & 24.50 & $+9.20$ & $[+6.50, +12.00]$ \\
\bottomrule
\end{tabular}

%% file: benchmark_rewrite_20260916/group_error_table.tex
\begin{tabular}{@{}lrrrrrr@{}}
\toprule
Source / configuration & 0 & 1 & 2 & 3 & 4 & 5 \\
\midrule
PlotQA / Qwen 3.8 Flash Next & 835 & 147 & 11 & 7 & 0 & 0 \\
PlotQA / Gemma 4 31B & 667 & 300 & 19 & 14 & 0 & 0 \\
PlotQA / Qwen 3.8 27B & 665 & 283 & 39 & 13 & 0 & 0 \\
PlotQA / Gemma 4 26B A4B & 539 & 419 & 22 & 16 & 4 & 0 \\
PlotQA / GLM 5.3 Flash & 273 & 558 & 167 & 1 & 1 & 0 \\
PlotQA / Molmo2-8B & 194 & 509 & 297 & 0 & 0 & 0 \\
CLEVR / Qwen 3.8 Flash Next & 454 & 506 & 28 & 11 & 1 & -- \\
CLEVR / Gemma 4 31B & 210 & 451 & 89 & 245 & 5 & -- \\
CLEVR / Qwen 3.8 27B & 273 & 505 & 144 & 69 & 9 & -- \\
CLEVR / Gemma 4 26B A4B & 189 & 491 & 79 & 228 & 13 & -- \\
CLEVR / GLM 5.3 Flash & 205 & 793 & 2 & 0 & 0 & -- \\
CLEVR / Molmo2-8B & 216 & 562 & 107 & 112 & 3 & -- \\
GQA / Qwen 3.8 Flash Next & 399 & 448 & 137 & 16 & -- & -- \\
GQA / Gemma 4 31B & 372 & 453 & 160 & 15 & -- & -- \\
GQA / Qwen 3.8 27B & 359 & 455 & 170 & 16 & -- & -- \\
GQA / Gemma 4 26B A4B & 392 & 400 & 192 & 16 & -- & -- \\
GQA / GLM 5.3 Flash & 305 & 651 & 42 & 2 & -- & -- \\
GQA / Molmo2-8B & 267 & 573 & 146 & 14 & -- & -- \\
\bottomrule
\end{tabular}

%% file: additional_appendix.tex
\section{Evidence Checks and Controlled Evaluations}
\label{app:additional}

This section documents the additional experiments and analyses in Sections~\ref{sec:extensions} and \ref{sec:additional}. Original benchmark results retain their original labels and denominators. Reviewed-subset scores and new controls have their own explicitly stated cohorts.

\subsection{Independent Scene Review and Paired Rescoring}
\label{app:scene-review}

\paragraph{Sampling and annotation.}
We sample 200 GQA and 100 CLEVR question groups independently of benchmark responses, excluding source images used in an earlier 20-group pilot. GQA includes 75 location and 125 other questions; CLEVR sampling is stratified by answer type, program depth, and hidden attribute. Reviewing every original state yields 600 GQA and 400 CLEVR views. Each annotator also reviews 418 control inputs: 172 answer-preserving occlusions, 178 color variants, and 68 inputs with replacement questions (Appendix~\ref{app:occlusion}). The total is 1,418 inputs per annotator.

GPT-6.1 Sol and Claude Opus 5.5 each receive one image and question in a fresh context, without the source target, state, other group views, benchmark responses, or the other annotator's response; the first successful structured judgment for each input is retained. The annotation prompt distinguishes \emph{answerable}, \emph{insufficient information}, \emph{invalid reference}, and \emph{uncertain}. It requests visible evidence; for insufficient information, it also requests the missing fact and two concretely supported alternatives. It explicitly cautions against inferring unanswerability from a patch, treating a cube as an occluder without visual support, guessing a hidden palette, or replacing an ambiguous referent with another object. The annotation schema also records reference issues and uncertainty.

\paragraph{Retention and correction.}
We compare these judgments with source answers, programs, and edit records. Supported views require both annotators to agree with the source answer after fixed answer normalization, without explicit uncertainty or reference flags. Unsupported views require agreement on an abstention category and consistency with source annotations and edit records. GQA masks intended to preserve the answer are flagged if they overlap a dependency box. Unmapped answers, disagreement, or source inconsistencies remain unresolved.

An AI assistant, with access to source labels and both annotations, additionally spot-checks contact sheets showing all views of 12 fixed random source groups (six per domain). These procedures define a source-consistent, model-reviewed subset, distinct from the executable certificates of Equation~\ref{eq:witness}. Retention rules were finalized during annotation and frozen before the new benchmark responses.

The spot audit identifies a GQA grounding error: a source node named ``ground'' with material ``wood'' is attached to a cistern, whereas the visible floor is tiled. Both annotators answer tile. All three reviewed views ($F,S,M$) are labeled answerable with target answer ``tile''; the original records remain available. This is one documented grounding correction affecting three inputs.

\begin{table}[ht]
\centering\small
\caption{\textbf{Original-state review coverage.} Retained and unresolved counts are views, not estimated label-error rates. Corrected views are included in Retained; the corrected GQA $M$ view is answerable.}
\label{tab:review-coverage}
\input{generated/review_coverage.tex}
\end{table}

Of the 1,000 original views, 479 are retained (183 GQA and 296 CLEVR) and 521 remain unresolved (Table~\ref{tab:review-coverage}). Judgment-category agreement is 74.5\% on GQA and 88.25\% on CLEVR (Cohen's $\kappa=0.411$ and $0.534$); agreement is not a measure of label validity. In particular, both annotators call 67 of the 200 GQA $M$ views answerable. These judgments flag residual evidence or source problems; agreement alone does not trigger a correction. Only 3 of 200 GQA groups and 21 of 100 CLEVR groups retain every original state. Consequently, the reviewed results below describe per-view task accuracy and the effect of the documented correction, rather than a new scene-wide ranking.

\begin{table}[ht]
\centering\small
\caption{\textbf{Six-model rescoring on the same retained inputs.} Task accuracy (\%) uses 183 GQA and 296 CLEVR views. $\Delta$ is reviewed-label accuracy minus original-label accuracy on the same inputs, in percentage points; every CLEVR $\Delta$ is zero. Brackets are source-image bootstrap 95\% intervals for reviewed accuracy.}
\label{tab:review-rescore}
\input{generated/review_rescore.tex}
\end{table}

The comparison in Table~\ref{tab:review-rescore} isolates relabeling from selection by comparing identical retained IDs before and after correction. Source-image bootstrap intervals use 10,000 replicates. The subset scores characterize the retained views; the original scene analyses continue to use the complete original cohorts.

\subsection{Occlusion, Appearance, and Question Controls}
\label{app:occlusion}

\paragraph{Construction and eligibility.}
For up to 100 groups per scene source, we construct an answer-preserving occluded view $S_{\rm occ}$. GQA uses the $M$ rectangle's width, height, color, and mask count at a position with no dependency-box overlap; 89 candidates satisfy the geometric constraints. CLEVR uses the same black matte cube replacement as $M$, selecting a source-program-preserving replacement; 83 candidates pass the program check. Two-model review retains 48 of 89 GQA control views and 80 of 83 CLEVR control views. Pairing each with a retained full view ($F$) having the same answer leaves 42 GQA and 74 CLEVR pairs. Requiring the original missing view ($M$) also to be reviewed as unanswerable leaves three GQA and 22 CLEVR three-view groups. Candidates are not replaced according to model outcomes.

Across the occlusion, color-change, and replacement-question controls, Qwen 3.8 Flash Next and Molmo2-8B each receive 818 inputs: 400 original $F/M$ views and the 418 new controls, including unresolved inputs; correctness analyses use only the pairs fixed before inference. Table~\ref{tab:occlusion-control} reports paired answer accuracy. GQA refusals are 4/42 on both $F$ and $S_{\rm occ}$ for Qwen, and 3/42 on both for Molmo2. On CLEVR, Qwen refusals increase from 3/74 to 4/74 and Molmo2 refusals from 21/74 to 22/74. Correct answers to occluded supported inputs directly contradict an unconditional ``occlusion implies abstention'' response rule on these inputs; the question-dependence control below addresses more selective appearance cues.

\begin{table}[ht]
\centering\small
\caption{\textbf{Accuracy with answer-preserving occlusion.} Correct answers on identical question groups; $S_{\rm occ}$ applies the missing-state edit to irrelevant content. $\Delta$ is occluded minus full-image accuracy, in percentage points, with a paired source-image bootstrap 95\% interval. Qwen denotes Qwen 3.8 Flash Next.}
\label{tab:occlusion-control}
\input{generated/occlusion_main.tex}
\end{table}

On these three-view groups, success requires correct answers on $F$ and $S_{\rm occ}$ and abstention on $M$. Qwen succeeds on 2/3 GQA and 18/22 CLEVR groups; Molmo2 succeeds on 0/3 GQA and 2/22 CLEVR groups. This three-condition control statistic is separate from original-state $J_k$.

\paragraph{Color and mask geometry.}
We change GQA mask color to near-black, RGB $(8,8,8)$, and CLEVR cube color to gray, preserving the source scene and rendering seed. GQA pixels outside the mask remain identical; CLEVR reflections can change. Review retains 69 of 178 color variants. Requiring the original and changed-color views to share the reviewed answerability label and, when answerable, target answer leaves 64 pairs: 17 GQA supported pairs, 36 CLEVR supported pairs, and 11 CLEVR unsupported pairs. Table~\ref{tab:appearance-control} reports all eligible comparisons. No GQA $M$ pair meets this criterion. The experiment tests the specified color changes, rather than natural cropping, blur, or arbitrary edits.

\begin{table}[ht]
\centering\small
\caption{\textbf{Color-variant task success.} Supported success requires the correct answer; unsupported success requires abstention. Counts compare original and changed color on the same IDs. $\Delta$ is changed-color minus original-color success, with its 95\% interval, in percentage points. Qwen denotes Qwen 3.8 Flash Next.}
\label{tab:appearance-control}
\input{generated/appearance.tex}
\end{table}

As a construction diagnostic, all 89 GQA candidate pairs match mask area, aspect ratio, and count exactly. To distinguish missing-state masks from answer-preserving controls using geometry, we fit logistic regression with L2 regularization to area, aspect ratio, count, and position. Five-fold cross-validation, keeping each source image within one fold, gives 50.6\% accuracy (95\% interval $[44.4,57.3]$). This assesses a linear geometric cue, not every possible visual shortcut.

\paragraph{Question dependence on the same image.}
For 40 groups, we attempt to construct a replacement question whose answer remains visible after masking. We obtain questions for 34 groups and review each on both the original and masked image, yielding 68 inputs. Review retains 51 individual views; requiring the same supported target on both leaves 8 GQA and 17 CLEVR pairs. Table~\ref{tab:question-control} reports answer accuracy. Each original/masked pair asks the same replacement question. Within the masked condition, the pixels are identical to those used for the original question. In five of the 17 CLEVR pairs, review also confirms that the original question is unanswerable. On the same masked image, Qwen correctly abstains on that question and answers the replacement question in all five cases; Molmo2 does so in two. No GQA group meets that switch criterion. These comparisons demonstrate question-dependent behavior on fixed occluded images; replacement questions are not difficulty matched.

\begin{table}[ht]
\centering\small
\caption{\textbf{Visible-target questions on original and masked images.} Correct answers for the same replacement question; every pair has a retained, identical target. Qwen denotes Qwen 3.8 Flash Next.}
\label{tab:question-control}
\input{generated/questions.tex}
\end{table}

All new scene contrasts use paired source-image bootstrap with 10,000 replicates. Intervals are descriptive and unadjusted for multiple comparisons. A zero-width empirical bootstrap interval means no variation in the observed paired differences, not population certainty. Invalid outputs count as failures; selection does not depend on benchmark responses.

\subsection{A Targeted Chart Localization Diagnostic}
\label{app:chart-control}

We fix 50 chart groups independently of model success and evaluate $F$ and $M$ with and without an x-position marker. Every image receives the same 28-pixel white header. Only the marked condition adds an arrow at the requested x position; it supplies neither the y value nor an indicator of the state. The chart body is unchanged. Both models (Qwen 3.8 Flash Next and Molmo2-8B) process all four conditions, producing 400 valid responses.

\begin{table}[ht]
\centering\small
\caption{\textbf{Chart localization control on 50 groups.} Entries are percentages, unmarked $\rightarrow$ marked. Joint success here requires a correct $F$ answer and $M$ abstention, and is distinct from $J_5$. Qwen denotes Qwen 3.8 Flash Next.}
\label{tab:chart-localization}
\input{generated/chart_control.tex}
\end{table}

Every $+2$-point change in Table~\ref{tab:chart-localization} has a paired source-image bootstrap 95\% interval of $[0,6]$ points (10,000 replicates). The marker does not reduce missing-state acceptance on these inputs. This tests whether the particular localization aid resolves the errors, without assigning each response to an internal mechanism.

For complementary qualitative evidence, a seeded sample of 12 valid archived chart-$M$ false acceptances per model is drawn from Qwen's 113 and GLM 5.3 Flash's 612 eligible responses. Eleven Qwen explanations claim a visible target label and one invokes a constant trend. Eleven GLM explanations claim a target label and one explicitly uses a different year; one label-claiming response also acknowledges the covered marker. An AI assistant codes these reported explanations into categories; they are not independently verified causal traces. Together with the intervention, they motivate separating localization, extrapolation, and instruction-following hypotheses, which Appendices~\ref{app:chart-instructions} and \ref{app:chart-new} test directly.

\subsection{What Joint Success Adds to State Marginals}
\label{app:joint-reference}

Observed joint success requires every obligation to hold for the same question. The independence reference instead asks what success would be if state outcomes were independent, holding each state's success rate fixed. Let $t_{gs}$ indicate task success for group $g$ and state $s$: a correct supported answer or a required abstention, with invalid outputs counted as failures. The original $J_k=G^{-1}\sum_g\prod_s t_{gs}$ preserves the grouping, whereas $J_{\rm ind}=\prod_s(G^{-1}\sum_g t_{gs})$ removes within-question association. Table~\ref{tab:joint-reference} gives both for each configuration and source. The conjunction is an established scoring rule; the empirical question is how often its obligations hold for the same question.

\begin{table}[ht]
\centering\small
\caption{\textbf{Observed joint success and independence reference.} Each configuration uses 1,000 groups per source. $J_{\rm ind}$ multiplies state-specific task success rates, including answer correctness. Scores are percentages; Difference is $J-J_{\rm ind}$ in percentage points.}
\label{tab:joint-reference}
\input{generated/joint_reference.tex}
\end{table}

For Qwen 3.8 Flash Next on GQA, task success is 70.9\% on full views, 68.8\% on supported edits, and 56.2\% on missing-state views. Their product is 27.41\%, versus observed $J=33.70\%$. Resampling complete groups and recomputing both quantities yields an excess of $6.29$ points with 95\% interval $[4.64,7.97]$. GLM 5.3 Flash's corresponding excess is $2.86$ points, $[1.43,4.27]$. Independently permuting state outcomes across groups while preserving each marginal gives central 95\% reference ranges of $[25.7,29.1]$ for Qwen and $[20.1,23.1]$ for GLM over 10,000 shuffles. Shuffle ranges describe the independence reference, not uncertainty in observed $J$.

Within-question association therefore contributes information beyond the marginals. Shared question difficulty and correlated errors can produce this association. A separate chart comparison shows how adding answer correctness reverses a decision-based ranking while keeping the groups fixed (Section~\ref{sec:ranking}).

\subsection{Inference Conditions and Interpretation}
\label{app:additional-inference}

New controls use the benchmark prompt in Appendix~\ref{app:protocol}, temperature zero, a 512-token output limit, schema-constrained JSON decoding, Qwen native thinking disabled, and bfloat16 (BF16) precision in vLLM. Paired conditions for a source image use the same model configuration. All 1,636 scene-control responses (818 inputs for each of two models) are recorded; one Qwen response reaches the token limit and is invalid. Its input was already unresolved before inference and is outside the eligible correctness cohorts. No outcome-based retry or exclusion is applied.

These controls characterize the specified configurations. Appendix~\ref{app:settings-new} adds a separate fixed-cohort comparison of prompt wording, generation budget, and native Qwen reasoning mode. Quantization is not separately ablated; Appendix~\ref{app:protocol} describes Qwen 3.8 27B, the only FP8 configuration, and its two serving deployments. The independent annotation models provide label-review evidence; they are not closed-model benchmark baselines.

\subsection{Executable CLEVR Completion Witnesses}
\label{app:clevr-witness}

For every CLEVR group, we construct two source worlds differing only in the hidden attribute and execute the published question program on both. All 1,000 pairs give different answers and identical masked renderer inputs. We then independently render both masked completions for the 100 groups already selected for the main review, using the original Blender 2.79b renderer and fixed rendering seed. Direct comparison of decoded pixel arrays gives exact equality for 100/100 pairs; both also match the released $M$ image. The 100 rendered pairs are not selected by inference outcomes; the remaining 900 pairs are verified structurally.

These witnesses instantiate the observation-equivalence condition of Equation~\ref{eq:witness} under the declared program, completion space, and renderer. They establish ambiguity within that task semantics. They do not establish that every natural-language referent is unambiguous or that a viewer necessarily interprets the black cube as an occluder. The answer-preserving black-cube controls in Appendix~\ref{app:occlusion} address the separate behavioral concern about appearance cues.

\subsection{Photographic Labels and Sensitivity to Missing-State Scoring}
\label{app:photo-sensitivity}

\paragraph{Agreement and source consistency answer different questions.}
The strict source-consistency filter in Appendix~\ref{app:scene-review} fixes cohorts for controlled comparisons. A complementary, post-hoc analysis reexamines the same 1,000 original views: 600 GQA and 400 CLEVR. It accepts answers shared by the two review models, including documented synonym and color-equivalence alignments, without requiring agreement with the original target. Of these 1,000 views, 719 have shared judgments: 579 agree with the original targets and 140 propose changes (35 answer changes, 11 supported-to-unsupported changes, and 94 unsupported-to-supported changes); 281 remain unresolved. These are model-consensus targets, not 140 certified source errors. Within the 200 GQA missing-state views, this analysis yields 64 shared answers, 13 shared abstention judgments, and 123 unresolved cases. We retain these strata separately rather than relabel the complete photographic benchmark.

\paragraph{Permissive scoring on all groups.}
For each of six models and all 1,000 GQA groups, let $t_F,t_S$ retain their original task-success definitions. At $M$, permissive success $t_M^+$ accepts a valid abstention or an answer compatible with the original full-image target; uncertain answer comparisons also receive credit. Then
\[
J^+ = \frac{1}{G}\sum_g t_{gF}t_{gS}t_{gM}^+,
\qquad
J_{-M} = \frac{1}{G}\sum_g t_{gF}t_{gS}.
\]
Invalid outputs remain failures. Under the assumption that the $F/S$ labels and answer references are correct and remain applicable to the masked question, $J^+$ is an optimistic bound allowing either $M$ response decision item by item. It is not one common, verified relabeling across models. $J_{-M}$ removes the $M$ obligation altogether. A third scenario makes every $M$ answerable and requires an answer compatible with the $F$ target, rather than crediting abstention. Table~\ref{tab:gqa-sensitivity} distinguishes these scenarios.

\begin{table}[ht]
\centering\small
\caption{\textbf{GQA joint success under four scoring rules (\%).} Every cell uses all 1,000 groups for the named model. Parentheses give ranks under original and permissive scoring. All alternative rules keep $F/S$ labels fixed. Permissive scoring credits ambiguous answer comparisons; it does not certify photographic labels.}
\label{tab:gqa-sensitivity}
\input{generated/gqa_sensitivity.tex}
\end{table}

Permissive $J$ ranges from 51.2\% to 63.5\%, leaving 36.5--48.8\% of groups incomplete under these favorable assumptions. Its substantial increase over the original score shows that the absolute failure rate depends on $M$ labels. Table~\ref{tab:gqa-sensitivity} reports both rankings: GLM 5.3 Flash moves from fifth to first, while Qwen 3.8 Flash Next moves from first to second. Under each of the three alternative GQA $M$ rules in Table~\ref{tab:gqa-sensitivity}, the mean three-source ordering changes only by exchanging Gemma 4 26B A4B and GLM 5.3 Flash in fourth and fifth place; the top three and the last position are unchanged. These are descriptive orderings, not significance tests.

The decision--task gap under permissive scoring is 18.5--29.7 points. Because either $M$ decision is accepted in that rule, this gap is not a lower bound on the original $B-J$ under every possible true relabeling. Likewise, disagreement between an $M$ answer and the $F$ reference is conditional evidence of error, not label-independent proof. The conditional $J^+$ bound is the appropriate sensitivity statement.

\begin{figure}[ht]
\centering
\includegraphics[width=.78\linewidth]{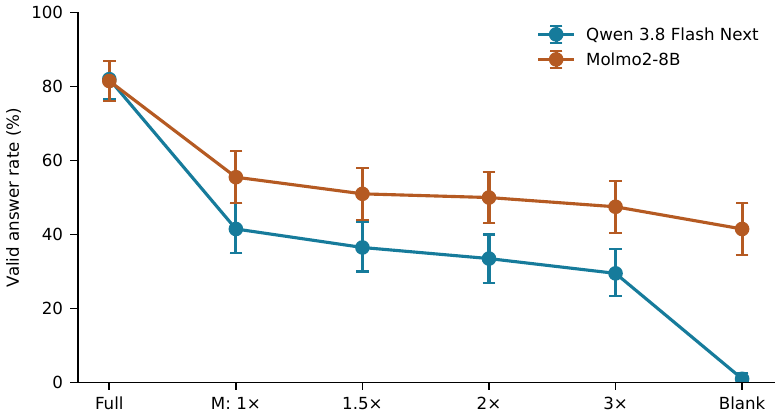}
\caption{\textbf{Answering as visible information is removed.} Each condition uses 200 questions per model; mask multipliers apply to width and height. Error bars are source-image bootstrap 95\% intervals. Fully blank inputs contain no scene pixels but retain the question's linguistic priors.}
\label{fig:mask-ladder}
\end{figure}

\paragraph{Nested masks and answer prediction.}
On the fixed 200 GQA groups, we enlarge the original $M$ rectangle's width and height by factors 1.5, 2, and 3, clipped at the image boundary, and add a completely uniform image. Enlargements are applied to the decoded original $M$; visible pixels outside each larger rectangle are unchanged. Duplicate observations caused by boundary clipping remain in the cohort. Intermediate masks are not assigned certified labels. Table~\ref{tab:mask-ladder} and Figure~\ref{fig:mask-ladder} report all 200 responses per condition and model.

\begin{table}[ht]
\centering\small
\caption{\textbf{Answer rates along the GQA mask ladder (\%).} A response counts as answering only if it satisfies the output contract. Every denominator is 200.}
\label{tab:mask-ladder}
\input{generated/mask_ladder.tex}
\end{table}

Answer rates decline with larger masks for both models. On fully blank images, Qwen 3.8 Flash Next answers 2/200 questions and Molmo2-8B answers 83/200. Answering therefore declines as visible information is removed, and Molmo2-8B additionally answers from question priors alone.

\begin{samepage}
Abstention selects different questions in each image condition, so we also require a guess on every input. Table~\ref{tab:guess-agreement} separates agreement among voluntarily answered inputs from agreement under this forced-guess prompt. For Qwen, the forced-guess match rate is 59.0\% on masked images versus 45.5\% on blank images: a 13.5-point difference, with paired 95\% interval $[5.0,22.0]$. This identifies predictive information remaining after masking, without establishing a unique visually determined answer.
\par\end{samepage}

\begin{table}[ht]
\centering\small
\caption{\textbf{GQA reference agreement by image condition and response rule.} Each condition contains 200 questions. Standard-prompt agreement is a percentage among answered inputs, with that denominator in parentheses. Forced-guess rates use all 200 inputs. String matching also credits emitted answers accompanied by an abstention declaration and is not task success. Qwen denotes Qwen 3.8 Flash Next; Molmo2 denotes Molmo2-8B.}
\label{tab:guess-agreement}
\begin{tabular}{@{}llrrr@{}}
\toprule
Model & Statistic & Full image & Original mask & Blank image \\
\midrule
Qwen & Standard-prompt agreement (\%) & 82.3 (164) & 77.1 (83) & 50.0 (2) \\
 & Forced-guess valid matches (\%) & 79.5 & 59.0 & 45.5 \\
 & Forced-guess string matches (\%) & 79.5 & 59.5 & 45.5 \\
\midrule
Molmo2 & Standard-prompt agreement (\%) & 84.7 (163) & 69.4 (111) & 43.4 (83) \\
 & Forced-guess valid matches (\%) & 62.5 & 27.5 & 0.0 \\
 & Forced-guess contract violations & 56/200 & 128/200 & 200/200 \\
 & Forced-guess string matches (\%) & 79.0 & 55.5 & 31.0 \\
\bottomrule
\end{tabular}
\end{table}

Molmo2 violates the forced-guess response contract on every blank image. Its emitted answer strings nevertheless match 31.0\% of references. Reporting these string matches separately distinguishes answer content from compliance with the required answerability decision.

\begin{figure}[ht]
\centering
\includegraphics[width=.42\linewidth]{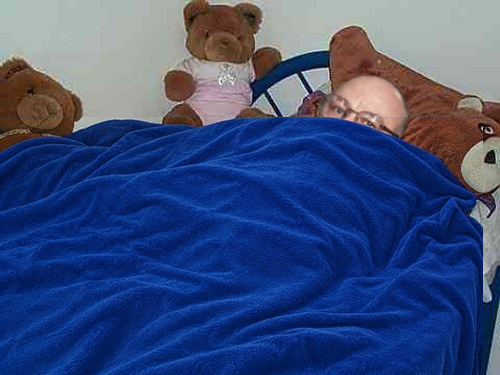}\hfill
\includegraphics[width=.42\linewidth]{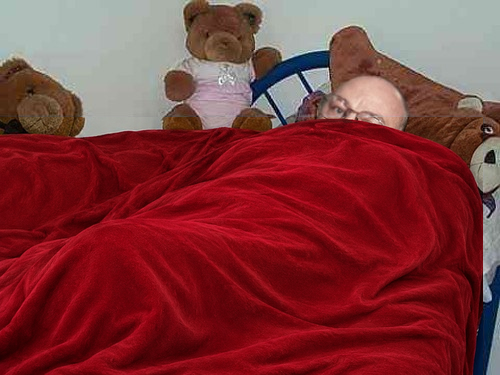}
\caption{\textbf{Alternative completions of a masked bedspread.} The question asks its color. Only pixels inside the original rectangle differ; generated content is illustrative and does not certify the full photographic cohort.}
\label{fig:photo-completions}
\end{figure}

\paragraph{Illustrative photographic completions.}
For one fixed bedspread-color question, two generated completions replace only the masked region with predominantly blue or red bedding (Figure~\ref{fig:photo-completions}). Both preserve every decoded pixel outside the mask. A visible boundary seam and texture changes remain; plausibility is assessed visually by an AI assistant. This is an illustrative pair, not a photographic certificate or a replay of the original JPEG encoding.

\begin{samepage}
\subsection{Chart Instructions, Visibility, and Numerical Compatibility}
\label{app:chart-instructions}

The same 50 chart groups used for localization are evaluated under the standard prompt and a uniqueness-and-examples prompt. The latter requires that visible information determine one answer, prohibits estimating obscured values, and contrasts a visible labeled value with a covered middle value compatible with multiple completions. It also warns against refusing because unrelated content is covered. Both $F$ and $M$ are evaluated for each model, totaling 400 responses. A separate visibility question asks whether the requested value is directly readable, adding 200 responses. Full prompt templates accompany the numerical evidence.
\par\end{samepage}

\begin{table}[ht]
\centering\small
\caption{\textbf{Chart instruction comparison (\%).} Standard $\to$ uniqueness-and-examples prompt on the same 50 groups. The interval is paired by source image; full-chart accuracy is reported to detect indiscriminate refusal.}
\label{tab:chart-instructions}
\input{generated/chart_instructions.tex}
\end{table}

Before these new responses, we fix a joint criterion: missing-state acceptance decreases by at least five percentage points with a paired 95\% interval excluding zero, and full-chart accuracy decreases by no more than five points. Molmo2 meets this criterion (Table~\ref{tab:chart-instructions}): $M$ acceptance changes from 46\% to 24\%, while $F$ accuracy changes from 38\% to 42\% (paired change interval $[0,10]$ points). Qwen's $M$ acceptance changes from 8\% to 2\%, with its interval including zero; $F$ accuracy remains 74\% (change interval $[-6,6]$). All 200 $M$ responses across models and prompts are format-valid. This contrast supports an instruction effect for Molmo2; it does not isolate whether the definition, examples, or explicit prohibitions cause the change.

Of 50 missing-state inputs, Qwen judges 11 values visible and 24 not visible; its remaining 15 responses lack a valid yes/no judgment. Molmo2 judges 12 visible and 38 not visible, with no invalid judgments. Molmo2 answers the original question despite separately judging its value not visible on 11/50 inputs; Qwen does so on 0/50.

We also analyze, post hoc, all 6,000 archived chart-$M$ responses. Of 2,212 answers, 1,239 (56.0\%) are numerically compatible with copying a neighboring visible value. Linear interpolation or extrapolation is compatible with 299 (13.5\%), including 277 that also match copying; only 22 (1.0\%) match a linear estimate without matching a copy candidate. The fixed tolerance is $|a-c|\leq\max(10^{-6},0.005\max(1,|c|))$. Without tolerance, 872/2,212 answers equal at least one candidate exactly. Numerical compatibility does not establish that a model used the corresponding operation.

\paragraph{Inference and uncertainty.}
The mask, guessing, visibility, and instruction conditions produce 4,200 responses (2,100 per model), separate from the earlier scene and localization controls. There are 3,794 task-format-valid responses; all 406 invalid responses remain in their denominators. Paired comparisons use 10,000 source-image bootstrap replicates. These descriptive intervals are not corrected for multiple comparisons. Models use vLLM, BF16, temperature zero, a 512-token limit, schema-constrained JSON decoding, and Qwen native thinking disabled; paired conditions use identical inference settings. Designs and criteria were fixed before these responses, after earlier results were known.

\subsection{Supplementary Model Audit of Photographic Masks}
\label{app:model-audit}

A further blinded audit revisits the GQA $M$ judgments from the earlier Sol--Claude review: all 64 shared-answer cases, all 13 shared-abstention cases, and a seeded sample of 40 unresolved cases. GPT-6.1 Sol and Gemini 4 Argon each judge all 117 images in fresh contexts without source answers, strata, or the other model's response. Both choose among answerable, insufficient information, invalid reference, and uncertain, under a prompt that differs from Appendix~\ref{app:scene-review}. Argon's reasoning setting and image interface are not uniform across its judgments. This audit does not alter benchmark labels or scoring.

\begin{table}[ht]
\centering\small
\caption{\textbf{Supplementary model judgments by original review stratum.} Agreement counts are descriptive; the 117 images are a deliberately stratified sample.}
\label{tab:model-audit}
\input{generated/supplementary_audit.tex}
\end{table}

Sol and Argon assign the same judgment category to 71/117 inputs (Table~\ref{tab:model-audit}) and disagree for 46. Both judge 33 answerable and 37 insufficient, with one shared invalid-reference judgment. Of the 33 jointly answerable inputs, 16 have matching normalized answer strings; differing strings are not automatically counted as wrong. Sol judges 70 answerable overall, versus 35 for Argon. Sol also participated in the earlier review, so this is not a fully independent new pair of model families. Because the sample is stratified, its agreement describes these strata rather than population label accuracy. It supports keeping photographic judgments distinct from executable certificates.

\input{revision_appendix.tex}

%% file: generated/review_coverage.tex
\begin{tabular}{@{}llrrrr@{}}
\toprule
Source & State & Reviewed & Retained & Unresolved & Corrected \\
\midrule
GQA & $F$ & 200 & 100 & 100 & 1 \\
GQA & $S$ & 200 & 69 & 131 & 1 \\
GQA & $M$ & 200 & 14 & 186 & 1 \\
CLEVR & $F$ & 100 & 91 & 9 & 0 \\
CLEVR & $S$ & 100 & 91 & 9 & 0 \\
CLEVR & $C$ & 100 & 87 & 13 & 0 \\
CLEVR & $M$ & 100 & 27 & 73 & 0 \\
\bottomrule
\end{tabular}

%% file: generated/review_rescore.tex
\begin{tabular}{@{}lrrr@{}}
\toprule
Model & GQA accuracy [95\%] & $\Delta$ & CLEVR accuracy [95\%] \\
\midrule
Qwen 3.8 Flash Next & 89.62 [83.68, 94.80] & $+1.64$ & 93.24 [89.27, 96.67] \\
Gemma 4 31B & 88.52 [82.39, 93.85] & $+1.64$ & 51.35 [43.34, 59.09] \\
Qwen 3.8 27B & 89.07 [83.42, 94.15] & $+1.64$ & 69.26 [62.70, 75.43] \\
Gemma 4 26B A4B & 85.25 [78.61, 91.26] & $+0.55$ & 44.26 [36.81, 51.85] \\
GLM 5.3 Flash & 97.81 [95.45, 99.48] & $+1.64$ & 80.07 [74.39, 85.52] \\
Molmo2-8B & 86.89 [80.87, 92.53] & $+1.64$ & 58.45 [51.06, 65.65] \\
\bottomrule
\end{tabular}

%% file: generated/occlusion_main.tex
\begin{tabular}{@{}llrrr@{}}
\toprule
Source & Model & $F$ & $S_{\rm occ}$ & $\Delta$ [95\% interval] \\
\midrule
GQA & Qwen & 38/42 & 38/42 & $+0.0\;[+0.0,+0.0]$ \\
GQA & Molmo2-8B & 39/42 & 39/42 & $+0.0\;[+0.0,+0.0]$ \\
CLEVR & Qwen & 67/74 & 65/74 & $-2.7\;[-9.5,+4.1]$ \\
CLEVR & Molmo2-8B & 38/74 & 36/74 & $-2.7\;[-10.8,+4.1]$ \\
\bottomrule
\end{tabular}

%% file: generated/appearance.tex
\begin{tabular}{@{}lllrrr@{}}
\toprule
Source & State & Model & Original & Changed & $\Delta$ [95\% interval] \\
\midrule
GQA & $S_{\rm occ}$ & Qwen & 15/17 & 15/17 & $+0.0\;[+0.0,+0.0]$ \\
GQA & $S_{\rm occ}$ & Molmo2-8B & 17/17 & 17/17 & $+0.0\;[+0.0,+0.0]$ \\
CLEVR & $S_{\rm occ}$ & Qwen & 30/36 & 32/36 & $+5.6\;[+0.0,+13.9]$ \\
CLEVR & $S_{\rm occ}$ & Molmo2-8B & 18/36 & 19/36 & $+2.8\;[-8.3,+13.9]$ \\
CLEVR & $M$ & Qwen & 10/11 & 11/11 & $+9.1\;[+0.0,+27.3]$ \\
CLEVR & $M$ & Molmo2-8B & 8/11 & 8/11 & $+0.0\;[-27.3,+27.3]$ \\
\bottomrule
\end{tabular}

%% file: generated/questions.tex
\begin{tabular}{@{}llrr@{}}
\toprule
Source & Model & Original image & Masked image \\
\midrule
GQA & Qwen & 8/8 & 8/8 \\
GQA & Molmo2-8B & 5/8 & 5/8 \\
CLEVR & Qwen & 17/17 & 17/17 \\
CLEVR & Molmo2-8B & 16/17 & 16/17 \\
\bottomrule
\end{tabular}

%% file: generated/chart_control.tex
\begin{tabular}{@{}lrrr@{}}
\toprule
Model & $F$ accuracy & $M$ acceptance & Two-condition joint \\
\midrule
Qwen & $74\rightarrow 76$ & $8\rightarrow 10$ & $72\rightarrow 74$ \\
Molmo2-8B & $38\rightarrow 40$ & $46\rightarrow 46$ & $26\rightarrow 28$ \\
\bottomrule
\end{tabular}

%% file: generated/joint_reference.tex
\begin{tabular}{@{}llrrr@{}}
\toprule
Source & Model & $J$ & $J_{\rm ind}$ & Difference \\
\midrule
PlotQA & Qwen 3.8 Flash Next & 57.00 & 48.23 & $+8.77$ \\
PlotQA & Gemma 4 31B & 55.60 & 45.01 & $+10.59$ \\
PlotQA & Qwen 3.8 27B & 34.10 & 25.46 & $+8.64$ \\
PlotQA & Gemma 4 26B A4B & 40.70 & 28.96 & $+11.74$ \\
PlotQA & GLM 5.3 Flash & 18.20 & 11.09 & $+7.11$ \\
PlotQA & Molmo2-8B & 6.10 & 3.64 & $+2.46$ \\
CLEVR & Qwen 3.8 Flash Next & 43.50 & 40.43 & $+3.07$ \\
CLEVR & Gemma 4 31B & 8.90 & 4.91 & $+3.99$ \\
CLEVR & Qwen 3.8 27B & 19.20 & 14.39 & $+4.81$ \\
CLEVR & Gemma 4 26B A4B & 5.40 & 2.82 & $+2.58$ \\
CLEVR & GLM 5.3 Flash & 17.00 & 11.69 & $+5.31$ \\
CLEVR & Molmo2-8B & 11.80 & 6.91 & $+4.89$ \\
GQA & Qwen 3.8 Flash Next & 33.70 & 27.41 & $+6.29$ \\
GQA & Gemma 4 31B & 28.90 & 22.75 & $+6.15$ \\
GQA & Qwen 3.8 27B & 28.80 & 23.76 & $+5.04$ \\
GQA & Gemma 4 26B A4B & 30.00 & 21.11 & $+8.89$ \\
GQA & GLM 5.3 Flash & 24.50 & 21.64 & $+2.86$ \\
GQA & Molmo2-8B & 22.30 & 20.11 & $+2.19$ \\
\bottomrule
\end{tabular}

%% file: generated/gqa_sensitivity.tex
\begin{tabular}{lrrrr}
\toprule
Model & Original & Permissive $M$ & Omit $M$ & Answerable $M$ \\
\midrule
Qwen 3.8 Flash Next & 33.7 (1) & 62.0 (2) & 67.0 & 28.3 \\
Gemma 4 31B & 28.9 (3) & 55.3 (4) & 61.8 & 26.4 \\
Qwen 3.8 27B & 28.8 (4) & 52.9 (5) & 57.8 & 24.1 \\
Gemma 4 26B A4B & 30.0 (2) & 51.2 (6) & 54.2 & 21.2 \\
GLM 5.3 Flash & 24.5 (5) & 63.5 (1) & 73.8 & 39.0 \\
Molmo2-8B & 22.3 (6) & 59.5 (3) & 66.8 & 37.2 \\
\bottomrule
\end{tabular}

%% file: generated/mask_ladder.tex
\begin{tabular}{lrrrrrr}
\toprule
Model & $F$ & $M$: $1\times$ & $1.5\times$ & $2\times$ & $3\times$ & Blank \\
\midrule
Qwen 3.8 Flash Next & 82.0 & 41.5 & 36.5 & 33.5 & 29.5 & 1.0 \\
Molmo2-8B & 81.5 & 55.5 & 51.0 & 50.0 & 47.5 & 41.5 \\
\bottomrule
\end{tabular}

%% file: generated/chart_instructions.tex
\begin{tabular}{llrl}
\toprule
Model & $M$ acceptance & $\Delta_M$ [95\% interval] & $F$ accuracy \\
\midrule
Qwen 3.8 Flash Next & $8.0\to2.0$ & $-$6.0 [$-$14.0, 0.0] & $74.0\to74.0$ \\
Molmo2-8B & $46.0\to24.0$ & $-$22.0 [$-$34.0, $-$12.0] & $38.0\to42.0$ \\
\bottomrule
\end{tabular}

%% file: generated/supplementary_audit.tex
\begin{tabular}{lrrrr}
\toprule
Earlier Sol--Claude judgment & $n$ & Same audit category & Both answerable & Both insufficient \\
\midrule
Shared answer & 64 & 39 & 32 & 7 \\
Shared abstention & 13 & 12 & 0 & 12 \\
Unresolved sample & 40 & 20 & 1 & 18 \\
\bottomrule
\end{tabular}

%% file: revision_appendix.tex
\subsection{Visible-Value Interventions on Certified Chart Inputs}
\label{app:chart-new}

Numerical compatibility alone cannot distinguish unsupported copying, interpolation, ordinary reading errors, or coincidence. We therefore intervene on visible values while preserving the question and the hidden target. This section distinguishes the earlier response-enriched diagnostic from a new disjoint follow-up.

\paragraph{Earlier diagnostic and its selection limitation.}
The earlier study samples 120 groups from the 822 groups on which at least one of six archived models supplied a numerical $M$ answer. It includes 68 targets with visible neighbors on both sides (interior), 46 with at least two visible neighbors on only one side (edge), and six with a single visible neighbor. Source-level edits hold axes, palette, question, and hidden target fixed. Clean-edit checks yield 73 nearest-neighbor shifts with corresponding $F$ controls, 33 second-neighbor edge edits, and 94 edits to another series at the queried x position. Each of the three models receives 513 inputs: 240 original full and missing-state views, 146 views from nearest-neighbor edits and their full-view controls, 33 second-neighbor edits, and 94 other-series edits. All 240 original rerenders match the released pixels. The 320 missing-state inputs (120 originals plus 200 edited views) each have two distinct hidden-value completions with identical pixels. These certificates use the fixed renderer with axis limits held constant.

A track requires numerical answers matching the same candidate prediction before and after the edit. Table~\ref{tab:earlier-tracks} reports tracking within tolerance $\max(10^{-6},0.005\max(1,|v|))$, where $v$ is the candidate value, and under exact rational equality. Zero observed tracks do not establish absence of the behavior.

\begin{table}[ht]
\centering\small
\caption{\textbf{Tracking visible values in the earlier, response-enriched cohort.} Each entry is matching pairs / eligible pairs. The restricted row retains only pairs answered correctly on both full-view controls. Gemma denotes Gemma 4 26B A4B; Qwen denotes Qwen 3.8 Flash Next.}
\label{tab:earlier-tracks}
\begin{tabular}{@{}lrrr@{}}
\toprule
Tracking criterion & Molmo2-8B & Gemma & Qwen \\
\midrule
Nearest neighbor, missing view & 8/73 & 5/73 & 1/73 \\
Nearest neighbor, full-view control & 0/73 & 0/73 & 0/73 \\
Missing view, both full controls correct & 5/54 & 4/63 & 0/68 \\
Nearest neighbor, exact match in $M$ & 2/73 & 4/73 & 1/73 \\
Second-neighbor extrapolation & 0/33 & 0/33 & 0/33 \\
Other series at queried position & 3/94 & 0/94 & 0/94 \\
\bottomrule
\end{tabular}
\end{table}

\begin{samepage}
The common neighbor shift changes both copying and linear-interpolation predictions, so it cannot separate those predictions by response movement alone. Also, the sample is enriched for previous numerical responding and cannot estimate prevalence across all charts. The next experiment addresses these two design limitations.
\par\end{samepage}

\paragraph{Disjoint sampling and prediction separation.}
\begin{samepage}
Before collecting new responses, we exclude the union of the earlier intervention cohort and the 50-group instruction/localization cohort: 165 distinct source images. We sample 240 of the remaining 835 chart groups with a fixed seed. Sampling reads source metadata and prior cohort IDs, not historical answers. All 240 original $F/M$ pairs remain in the baseline analysis. These are dot-line charts from the original held-out rendering family. Because the earlier cohort used response-based selection, exclusion of its images can change the remaining population: the new sample characterizes this disjoint remainder, not an unbiased estimate for all 1,000 charts.
\par\end{samepage}

For an interior target at $x_t$ between $x_l$ and $x_r$, define
\[
\alpha=\frac{x_r-x_t}{x_r-x_l},\qquad
\beta=\frac{x_t-x_l}{x_r-x_l},\qquad
L=\alpha y_l+\beta y_r.
\]
Categorical axes use their displayed index positions. We set $y'_l=y_l+\delta$ and $y'_r=y_r-(\alpha/\beta)\delta$. Thus the interpolant $L$ is exactly unchanged while each neighbor moves. For equal spacing, $10,30\to15,25$ illustrates the contrast. The target value remains fixed in a matching visible-target $F$ control. The model receives each image and question independently, without a state label or paired image.

The deterministic editing rule tries signed changes of 10\%, 7.5\%, 5\%, and 12.5\% of the original fixed y-axis span, alternating the first sign by group index. It requires in-axis values, nonoverlapping old/new neighbor and interpolation predictions under the scoring tolerance, unchanged unrelated drawing operations, and no changed pixels outside the edited markers, labels, and adjacent line segments. Of 240 sampled groups, 103 lack an interior target. Among the remaining 137, 63 admit no edit satisfying these constraints, leaving 74 edits with both $F$ and $M$ controls; no excluded group is replaced. These checks precede inference.

\begin{figure}[ht]
\centering
\begin{tikzpicture}
\node[inner sep=0pt, anchor=south west] (bal) at (0,0) {\includegraphics[width=\linewidth]{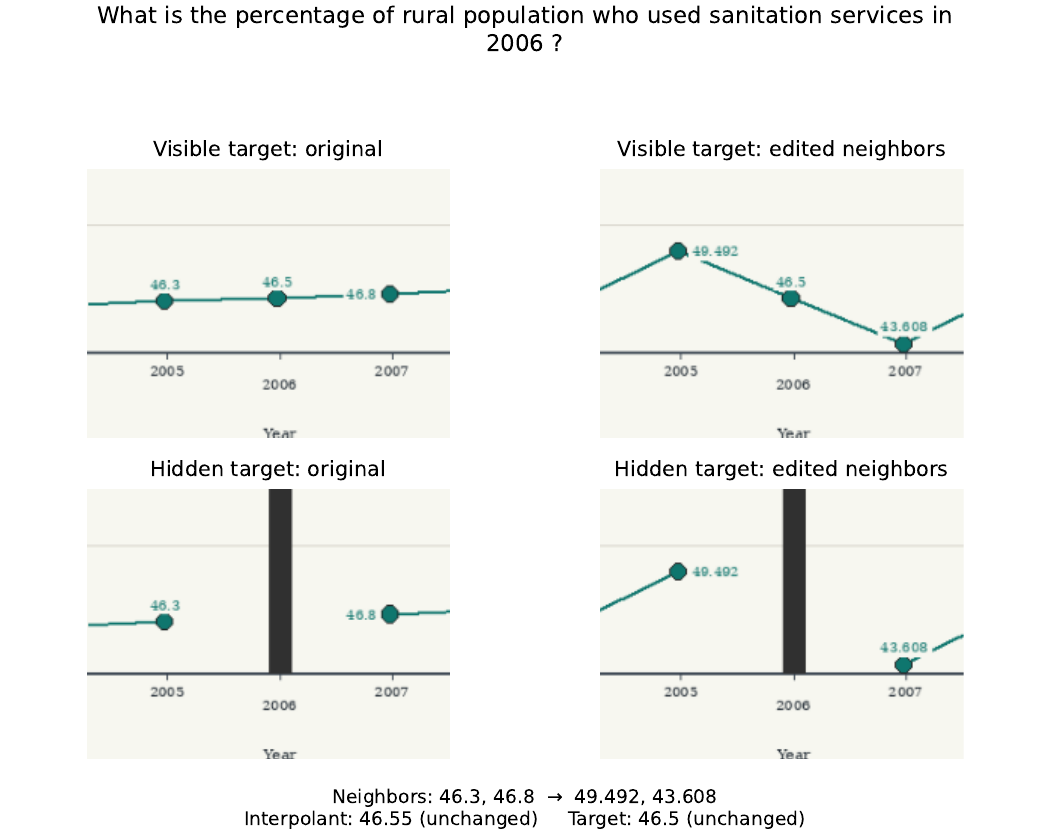}};
\begin{scope}[x={(bal.south east)}, y={(bal.north west)}]
\fill[white] (0.075,0.474) rectangle (0.925,0.494);
\fill[white] (0.075,0.093) rectangle (0.925,0.113);
\end{scope}
\end{tikzpicture}
\caption{\textbf{Separating neighbor-value movement from a fixed interpolation prediction.} Identical-coordinate crops of the queried series in an eligible group selected for readable labels before response inspection. The left and right neighbors change from (46.3, 46.8) to (49.492, 43.608) while the interpolant stays 46.55 and the target stays 46.5. The visible-target controls preserve the answer; the missing-target inputs retain their ambiguity witnesses. Models receive the complete images independently.}
\label{fig:balanced-inputs}
\end{figure}

Figure~\ref{fig:balanced-inputs} illustrates one edit. Every one of the 314 $M$ images (240 baselines plus 74 edits) has two distinct hidden-target values that produce identical decoded pixel arrays. Original $F/M$ rerenders match all 480 released images, and every evaluated input is verified to be pixel-identical to its rendering. We flag targets equal to the minimum or maximum value across all series in the original chart and separately report results excluding these groups.

\paragraph{Estimands and scoring.}
A neighbor track requires a numerical answer matching the same neighbor before and after its edit. The denominator is all 74 eligible pairs, including refusal and invalid output (Table~\ref{tab:balanced-neighbors}). We report the matching $F$ rate, paired $M-F$ difference, $F$ accuracy change, the subset correct on both $F$ controls, exact-rational matching, and exclusion of original target extrema. Compatibility with the unchanged interpolant is a separate descriptive count. The fixed numerical extractor is retained from the earlier intervention; tolerance and exact counts are both reported. The extractor requires exactly one numeric substring. It can reject units containing digits: the ``2'' in CO2 is treated as an additional number.

\input{generated/confirmation_results.tex}

\paragraph{Uncertainty and interpretation.}
Paired differences use 10,000 source-image bootstrap draws. Tracking proportions additionally receive Wilson intervals so a zero count is not represented as zero population uncertainty. Intervals are descriptive and unadjusted for multiple comparisons. The intervention tests dependence of outputs on visible values under certified missing evidence; it does not establish that every matching response was generated by copying, or that missing-state behavior has a single cause.

\subsection{Prompt, Native Reasoning, and Generation-Budget Comparison}
\label{app:settings-new}

\paragraph{Fixed cohorts and response obligations.}
We sample 60 chart groups from the new 240-group cohort and 60 CLEVR groups from the full source cohort, without consulting responses. All original states are included: 300 chart views ($F,S,C,M,I$) and 240 CLEVR views ($F,S,C,M$). Each source has 60 distinct source images. The samples are fixed before inference, and no input is replaced after a malformed or truncated response. Unlike an $F/M$ pair score, $J$ here uses every original benchmark state.

\paragraph{Independent factors.}
Qwen 3.8 Flash Next is evaluated in a $2\times2$ factorial contrast: native thinking disabled/enabled and maximum generation length 512/2,048 tokens. Both budgets apply to the full generated completion, including native thinking tokens. We verify that the rendered chat templates differ under the native toggle. Enabling this checkpoint's native mode also inserts its built-in system instruction requesting \texttt{xhigh} reasoning effort. Molmo2-8B receives the same two output limits; its template has no corresponding native thinking switch, so no such contrast is claimed. Both checkpoints additionally receive the exact uniqueness-and-examples prompt from Appendix~\ref{app:chart-instructions} on the 300 chart views at 512 tokens with thinking disabled. We do not apply chart-specific examples to CLEVR.

These comparisons request JSON output without constraining decoding to a JSON grammar. Primary scoring requires the entire final segment after native reasoning to be one JSON object; a surrounding code fence is allowed. Each contrast uses the same output rule and decoder. This permits native reasoning without a grammar that might prevent it.

The other benchmark-model follow-ups use schema-constrained JSON decoding (Appendices~\ref{app:occlusion}, \ref{app:chart-control}, \ref{app:photo-sensitivity}, \ref{app:chart-instructions}, and \ref{app:chart-new}). We therefore collect a fresh unconstrained baseline. Table~\ref{tab:settings-bridge-full} compares each setting with archived benchmark responses on the same groups. Settings responses are analyzed separately from the visible-value intervention because their output constraints differ.

\paragraph{Scoring and uncertainty.}
Primary supported-answer accuracy and $J$ use the released exact scoring rules: the original numeric parser and rational equality for charts, and rational equality or normalized exact text for CLEVR. The chart parser accepts commas, limited currency prefixes, fractional notation, and thousand/million/billion suffixes; percent signs do not rescale the number. We report supported accuracy, correct $M/I$ abstention, all-state validity, $B$, $J$, invalid outputs, and truncations (Tables~\ref{tab:inference-settings} and \ref{tab:settings-paired}; Figure~\ref{fig:settings-new}). Invalid outputs are failures rather than abstentions. Paired differences retain all 60 source-image clusters, using 10,000 bootstrap draws. Each arm uses one first-response pass; intervals resample source images and do not quantify variability across repeated serving runs.

\input{generated/settings_results.tex}

\paragraph{Runtime.}
The intervention and settings runs use vLLM in BF16 with temperature 0 and top-p 1, and paired conditions share every nonvaried setting.

\subsection{Decision--Task Reversal, Scene Question Types, and Archived-Protocol Comparison}
\label{app:new-analyses}

These post-hoc analyses rescore existing responses; they require no new inference. Paired intervals use 10,000 bootstrap draws over question groups.

\begin{samepage}
\paragraph{Decision and task metrics order two chart configurations oppositely.}
Table~\ref{tab:chart-reversal} compares Gemma 4 26B A4B and Qwen 3.8 27B on the 1,000 certified chart groups. Per-view decision failure and complete decision success favor Qwen 3.8 27B, whereas complete task success favors Gemma 4 26B A4B, because Qwen 3.8 27B returns wrong supported values in 324/1,000 groups whose decisions are all correct, compared with 132/1,000 for Gemma.
\par\end{samepage}

\begin{table}[ht]
\centering\small
\caption{\textbf{Opposite orderings on certified charts.} $E$ is per-view decision failure, $B_5$ complete decision success, and $J_5$ complete task success (all percentages). The last column counts groups with every decision correct but a wrong supported answer. Differences are Gemma 4 26B A4B minus Qwen 3.8 27B, in percentage points, with paired 95\% intervals.}
\label{tab:chart-reversal}
\footnotesize
\setlength{\tabcolsep}{3pt}
\begin{tabular}{@{}lrrrr@{}}
\toprule
Configuration & $E\downarrow$ & $B_5\uparrow$ & $J_5\uparrow$ & \shortstack{Correct decisions,\\wrong answer} \\
\midrule
Qwen 3.8 27B & 8.00 & 66.5 & 34.1 & 324 \\
Gemma 4 26B A4B & 10.54 & 53.9 & 40.7 & 132 \\
\midrule
Difference & $+2.54\;[+1.66,+3.46]$ & $-12.6\;[-16.3,-9.0]$ & $+6.6\;[+2.9,+10.3]$ & \\
\bottomrule
\end{tabular}
\end{table}

\paragraph{CLEVR acceptance by answer type.}
\begin{samepage}
Table~\ref{tab:clevr-types} partitions the 1,000 CLEVR $M$ views by the source answer type. Counting questions elicit the most acceptance for five of six configurations. For Qwen 3.8 Flash Next, supported-state refusal is 1.0\% for count questions, 6.1\% for yes/no questions, and 4.2\% for attribute questions, and GLM 5.3 Flash refuses at most 0.2\% of supported views of any type; for these configurations the contrast is specific to missing evidence rather than a general reluctance to answer one question type. Both review models, GPT-6.1 Sol and Claude Opus 5.5, judge 17/27 reviewed count $M$ views answerable, compared with 4/41 attribute views, most often by counting the black cube as a scene object (Appendix~\ref{app:scene-review}); these review judgments identify a literal reading compatible with count answers.
\par\end{samepage}

\begin{table}[ht]
\centering\small
\caption{\textbf{CLEVR missing-state acceptance by answer type.} Valid responses that declare an $M$ view answerable, with percentages of each type's groups. The occluder is identical across types.}
\label{tab:clevr-types}
\begin{tabular}{@{}lrrr@{}}
\toprule
Configuration & Count ($n=265$) & Yes/no ($n=327$) & Attribute ($n=408$) \\
\midrule
Qwen 3.8 Flash Next & 228 (86.0) & 161 (49.2) & 89 (21.8) \\
Gemma 4 31B & 165 (62.3) & 183 (56.0) & 87 (21.3) \\
Qwen 3.8 27B & 200 (75.5) & 204 (62.4) & 121 (29.7) \\
Gemma 4 26B A4B & 180 (67.9) & 187 (57.2) & 127 (31.1) \\
GLM 5.3 Flash & 239 (90.2) & 265 (81.0) & 291 (71.3) \\
Molmo2-8B & 123 (46.4) & 152 (46.5) & 230 (56.4) \\
\bottomrule
\end{tabular}
\end{table}

\paragraph{Every inference-setting arm against the archived protocol.}
Table~\ref{tab:settings-bridge-full} compares each arm in Appendix~\ref{app:settings-new} with the archived benchmark responses on the same 60 groups. The unconstrained standard-prompt nonthinking arms reproduce the archived Molmo2-8B scores exactly and the archived Qwen 3.8 Flash Next CLEVR score within 3.3 points. Final-JSON extraction recovers most of Qwen's chart deficit without schema-constrained decoding (Table~\ref{tab:format-sensitivity}). With native reasoning and 2,048 tokens, Qwen scores 56.7\% on charts and 45.0\% on CLEVR, compared with archived scores of 58.3\% and 43.3\%, respectively.

\begin{table}[ht]
\centering\small
\caption{\textbf{Inference arms versus the archived benchmark protocol ($J$, \%).} Each row uses the same 60 groups and all original states. $\Delta$ is the arm minus the archived score, with a paired 95\% interval in percentage points.}
\label{tab:settings-bridge-full}
\begin{tabular}{@{}lllrrr@{}}
\toprule
Model & Source & Setting & Archived $J$ & Setting $J$ & $\Delta$ [95\% interval] \\
\midrule
Qwen 3.8 Flash Next & Chart & Off / 512 & 58.3 & 33.3 & $-25.0\;[-38.3,-13.3]$ \\
 & & Off / 2,048 & 58.3 & 30.0 & $-28.3\;[-41.7,-15.0]$ \\
 & & On / 512 & 58.3 & 53.3 & $-5.0\;[-15.0,+5.0]$ \\
 & & On / 2,048 & 58.3 & 56.7 & $-1.7\;[-11.7,+10.0]$ \\
 & & Unique / 512 & 58.3 & 0.0 & $-58.3\;[-70.0,-46.7]$ \\
Qwen 3.8 Flash Next & CLEVR & Off / 512 & 43.3 & 40.0 & $-3.3\;[-10.0,+3.3]$ \\
 & & Off / 2,048 & 43.3 & 45.0 & $+1.7\;[+0.0,+5.0]$ \\
 & & On / 512 & 43.3 & 35.0 & $-8.3\;[-18.3,+1.7]$ \\
 & & On / 2,048 & 43.3 & 45.0 & $+1.7\;[-8.3,+11.7]$ \\
Molmo2-8B & Chart & Off / 512 & 13.3 & 13.3 & $0.0\;[0.0,0.0]$ \\
 & & Off / 2,048 & 13.3 & 13.3 & $0.0\;[0.0,0.0]$ \\
 & & Unique / 512 & 13.3 & 11.7 & $-1.7\;[-8.3,+3.3]$ \\
Molmo2-8B & CLEVR & Off / 512 & 15.0 & 15.0 & $0.0\;[0.0,0.0]$ \\
 & & Off / 2,048 & 15.0 & 15.0 & $0.0\;[0.0,0.0]$ \\
\bottomrule
\end{tabular}
\end{table}

%% file: generated/confirmation_results.tex
\begin{table}[ht]
\centering\small
\caption{\textbf{Balanced-neighbor intervention.} All 74 eligible pairs, sampled without consulting past answers. Tracks require matching the same neighbor before and after its edit; the interpolant stays fixed. $M-F$ subtracts full-view from missing-state tracking, in percentage points. $F$ accuracy uses the declared 0.5\% diagnostic tolerance. Parentheses give exact-rational $M$ tracks. Gemma 26B: Gemma 4 26B A4B; Qwen: Qwen 3.8 Flash Next.}
\label{tab:balanced-neighbors}
\begin{tabular}{@{}lrrrr@{}}\toprule
Model & $M$ tracks & $F$ tracks & $F$ correct, before/after & $M-F$ [95\% interval]\\\midrule
Molmo2 & 3/74 (0) & 0/74 & 63/74 $\to$ 64/74 & $+$4.1 [$+$0.0, $+$9.5] \\
Gemma 26B & 11/74 (9) & 2/74 & 67/74 $\to$ 66/74 & $+$12.2 [$+$4.1, $+$20.3] \\
Qwen & 0/74 (0) & 1/74 & 73/74 $\to$ 69/74 & $-$1.4 [$-$4.1, $+$0.0] \\
\bottomrule\end{tabular}\end{table}
\begin{samepage}
All 1,884 intervention responses are retained: three models each receive 628 inputs (480 original views and 148 edited views). Molmo2-8B has 120/240 baseline $M$ acceptances; Gemma 4 26B A4B has 104/240 baseline $M$ acceptances; Qwen 3.8 Flash Next has 29/240 baseline $M$ acceptances. Among pairs answered correctly on both full-view controls, missing-state tracking occurs in 2/59 pairs for Molmo2, 7/63 for Gemma 26B, and 0/68 for Qwen. Excluding targets at the original chart's global minimum or maximum yields 3/70 for Molmo2, 11/70 for Gemma 26B, 0/70 for Qwen.
\par\end{samepage}

The unchanged interpolant matches both $M$ responses in none of the 74 eligible pairs for any model; the pairs with numerical answers on both $M$ inputs number 12 for Molmo2, 23 for Gemma 26B, and 1 for Qwen. The 95\% Wilson intervals for $M$ tracking are Molmo2 [1.4, 11.3]\%, Gemma 26B [8.5, 24.7]\%, Qwen [0.0, 4.9]\%.

%% file: generated/settings_results.tex
\begin{table}[ht]
\centering\small
\caption{\textbf{Fixed-cohort inference comparison (\%).} Each domain has 60 groups; supported accuracy $A$ uses 180 views, $M$ abstention 60, and chart $I$ abstention 60. $V$ requires valid outputs on every state, $B$ every correct decision, and $J$ every correct task response. On/Off enable/disable native reasoning under the standard prompt; Unique uses chart-specific examples with reasoning off. The token limit includes reasoning and the final answer. Invalid/truncated counts are over 300 chart or 240 CLEVR responses; truncations are included in the invalid counts. All settings use unconstrained final-JSON decoding. Qwen: Qwen 3.8 Flash Next; Molmo2: Molmo2-8B.}
\label{tab:inference-settings}
\setlength{\tabcolsep}{3pt}
\begin{tabular}{@{}lllrrrrrrr@{}}\toprule
Model & Source & Mode / tokens & $A$ & $M$ & $I$ & $V$ & $B$ & $J$ & \shortstack{Invalid/\\trunc.}\\\midrule
Qwen & Chart & Off / 512 & 81.1 & 78.3 & 65.0 & 66.7 & 53.3 & 33.3 & 23/0 \\
Qwen & Chart & Off / 2,048 & 80.0 & 73.3 & 65.0 & 65.0 & 48.3 & 30.0 & 26/0 \\
Qwen & Chart & On / 512 & 81.1 & 81.7 & 90.0 & 83.3 & 71.7 & 53.3 & 12/12 \\
Qwen & Chart & On / 2,048 & 83.3 & 86.7 & 88.3 & 96.7 & 71.7 & 56.7 & 2/2 \\
Qwen & Chart & Unique / 512 & 58.3 & 20.0 & 1.7 & 1.7 & 0.0 & 0.0 & 153/0 \\
\midrule
Qwen & CLEVR & Off / 512 & 93.3 & 45.0 & -- & 95.0 & 43.3 & 40.0 & 4/0 \\
Qwen & CLEVR & Off / 2,048 & 92.2 & 48.3 & -- & 91.7 & 46.7 & 45.0 & 7/0 \\
Qwen & CLEVR & On / 512 & 83.3 & 40.0 & -- & 61.7 & 35.0 & 35.0 & 43/42 \\
Qwen & CLEVR & On / 2,048 & 94.4 & 46.7 & -- & 90.0 & 45.0 & 45.0 & 8/8 \\
\midrule
Molmo2 & Chart & Off / 512 & 57.8 & 56.7 & 56.7 & 100.0 & 31.7 & 13.3 & 0/0 \\
Molmo2 & Chart & Off / 2,048 & 57.8 & 56.7 & 56.7 & 100.0 & 31.7 & 13.3 & 0/0 \\
Molmo2 & Chart & Unique / 512 & 60.0 & 70.0 & 35.0 & 100.0 & 25.0 & 11.7 & 0/0 \\
\midrule
Molmo2 & CLEVR & Off / 512 & 54.4 & 38.3 & -- & 98.3 & 20.0 & 15.0 & 4/0 \\
Molmo2 & CLEVR & Off / 2,048 & 54.4 & 38.3 & -- & 98.3 & 20.0 & 15.0 & 4/0 \\
\bottomrule\end{tabular}\end{table}
\begin{table}[ht]
\centering\small
\caption{\textbf{Paired changes in full-group success.} Each contrast holds the cohort, checkpoint, and nonvaried decoder settings fixed. $\Delta J$ is first arm minus second, in percentage points; intervals use 10,000 paired source-image draws.}
\label{tab:settings-paired}
\begin{tabular}{@{}lllrr@{}}\toprule
Model & Source & Contrast & $\Delta J$ & 95\% interval\\\midrule
Qwen & Chart & Off / 2,048 minus Off / 512 & $-$3.3 & [$-$10.0, $+$3.3] \\
Qwen & CLEVR & Off / 2,048 minus Off / 512 & $+$5.0 & [$+$0.0, $+$11.7] \\
Qwen & Chart & Unique / 512 minus Off / 512 & $-$33.3 & [$-$45.0, $-$21.7] \\
Qwen & Chart & On / 512 minus Off / 512 & $+$20.0 & [$+$8.3, $+$31.7] \\
Qwen & CLEVR & On / 512 minus Off / 512 & $-$5.0 & [$-$16.7, $+$5.0] \\
Qwen & Chart & On / 2,048 minus Off / 2,048 & $+$26.7 & [$+$13.3, $+$40.0] \\
Qwen & CLEVR & On / 2,048 minus Off / 2,048 & $+$0.0 & [$-$10.0, $+$10.0] \\
Qwen & Chart & On / 2,048 minus On / 512 & $+$3.3 & [$-$5.0, $+$11.7] \\
Qwen & CLEVR & On / 2,048 minus On / 512 & $+$10.0 & [$+$1.7, $+$20.0] \\
Molmo2 & Chart & Off / 2,048 minus Off / 512 & $+$0.0 & [$+$0.0, $+$0.0] \\
Molmo2 & CLEVR & Off / 2,048 minus Off / 512 & $+$0.0 & [$+$0.0, $+$0.0] \\
Molmo2 & Chart & Unique / 512 minus Off / 512 & $-$1.7 & [$-$8.3, $+$3.3] \\
\bottomrule\end{tabular}\end{table}
All 3,840 planned setting responses are retained. Qwen contributes 2,460 responses, including 278 invalid and 64 truncated outputs; Molmo2 contributes 1,380 responses, including 8 invalid and 0 truncated outputs. Truncated completions are generation-budget failures, not abstentions.

Neither nonthinking Qwen arm reaches its token ceiling, so the generation budget binds only with native reasoning; the Off/512 versus Off/2,048 differences in Table~\ref{tab:settings-paired} therefore reflect run-to-run serving variation rather than the budget.
\begin{samepage}
\paragraph{Post-hoc format sensitivity.}
Inspection of completed arms revealed nonthinking responses that prepend prose to an otherwise valid final JSON. We therefore additionally extract exactly one schema-valid JSON object from the final segment after any thinking delimiter. Multiple valid objects, plain-text answers, and truncated completions remain failures. This post-hoc diagnostic changes only extraction; the primary scores and original responses are retained. Table~\ref{tab:format-sensitivity} distinguishes output-contract compliance from the answers recoverable under this more permissive parser.
\par\end{samepage}

\begin{table}[ht]
\centering\small
\caption{\textbf{Post-hoc final-JSON extraction.} Recovered/invalid counts outputs with one recoverable JSON object out of all initially invalid outputs (a dash means none were invalid); every domain still contains all 60 groups. $J$ is the primary strict score and $J_{\rm JSON}$ uses the diagnostic parser. Truncations remain failures under both rules.}
\label{tab:format-sensitivity}
\begin{tabular}{@{}lllrrr@{}}\toprule
Model & Source & Mode / tokens & Recovered/invalid & $J$ (\%) & $J_{\rm JSON}$ (\%)\\\midrule
Qwen & Chart & Off / 512 & 22/23 & 33.3 & 55.0 \\
Qwen & Chart & Off / 2,048 & 26/26 & 30.0 & 55.0 \\
Qwen & Chart & On / 512 & 0/12 & 53.3 & 53.3 \\
Qwen & Chart & On / 2,048 & 0/2 & 56.7 & 56.7 \\
Qwen & Chart & Unique / 512 & 94/153 & 0.0 & 18.3 \\
Qwen & CLEVR & Off / 512 & 0/4 & 40.0 & 40.0 \\
Qwen & CLEVR & Off / 2,048 & 0/7 & 45.0 & 45.0 \\
Qwen & CLEVR & On / 512 & 1/43 & 35.0 & 36.7 \\
Qwen & CLEVR & On / 2,048 & 0/8 & 45.0 & 45.0 \\
Molmo2 & Chart & Off / 512 & -- & 13.3 & 13.3 \\
Molmo2 & Chart & Off / 2,048 & -- & 13.3 & 13.3 \\
Molmo2 & Chart & Unique / 512 & -- & 11.7 & 11.7 \\
Molmo2 & CLEVR & Off / 512 & 0/4 & 15.0 & 15.0 \\
Molmo2 & CLEVR & Off / 2,048 & 0/4 & 15.0 & 15.0 \\
\bottomrule\end{tabular}\end{table}
\begin{samepage}
Qwen's uniqueness arm produces 153/300 invalid outputs, with primary $J=0.0\%$. The diagnostic recovers 94 outputs and yields $J_{\rm JSON}=18.3\%$; remaining format failures prevent interpreting this score as a pure change in visual reasoning. Molmo2's chart arms contain no invalid responses, so its $M/I$ trade-off is present under both parsers.
\par\end{samepage}
\begin{figure}[ht]
\centering
\includegraphics[width=\linewidth]{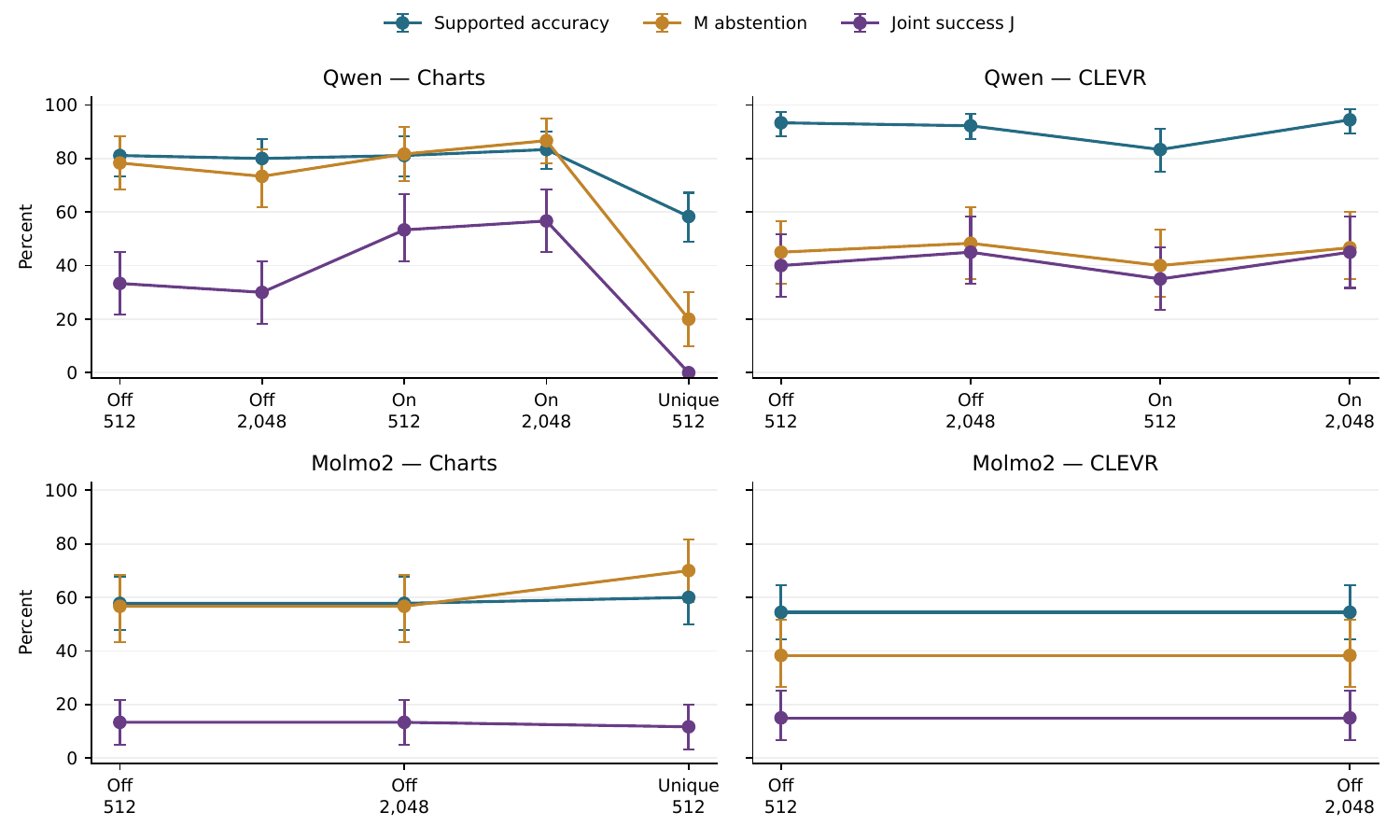}
\caption{\textbf{Supported accuracy, $M$ abstention, and joint success must be read together.} Each source uses the same 60 groups and every original state. Off/On disable/enable native reasoning; Unique denotes the examples prompt; numbers give token limits. Lines connect conditions for readability and do not imply a continuous treatment. Error bars show descriptive source-image bootstrap intervals.}
\label{fig:settings-new}
\end{figure}